\documentclass[11pt]{article}
\usepackage[a4paper,margin=25mm]{geometry}
\usepackage[T1]{fontenc}
\usepackage[utf8]{inputenc}
\usepackage{newtxtext,newtxmath}
\usepackage{amsmath}
\usepackage{graphicx}
\usepackage{booktabs,tabularx,longtable,array}
\usepackage{caption}
\usepackage{float}
\usepackage[section]{placeins}
\usepackage{fancyvrb}
\usepackage{xcolor}
\usepackage[hidelinks]{hyperref}
\usepackage{xurl}
\usepackage{microtype}
\usepackage{setspace}
\usepackage{indentfirst}
\DefineVerbatimEnvironment{CodeBlock}{Verbatim}{fontsize=\small,frame=single,framesep=3mm,rulecolor=\color{black!35},xleftmargin=0.5em,xrightmargin=0.5em}

\makeatletter
\newcommand{\plainfootnotetext}[1]{\begingroup\renewcommand{\thefootnote}{}\renewcommand{\@makefntext}[1]{\noindent ##1}\footnotetext{#1}\endgroup}
\makeatother
\title{High-dimensional state trajectories reveal transformer inference dynamics}
\author{Heng Zhao$^{1,*}$, Yufei Wang$^{2}$, Ruidian Song$^{3}$, Weisheng Chen$^{4}$\\[0.6em]
\small $^{1}$Independent Researcher, Beijing 100071, China\\
\small $^{2}$China Mobile Communications Group Co., Ltd., Beijing 100054, China\\
\small $^{3}$Sinotrans Logistics Co., Ltd., Beijing 100054, China\\
\small $^{4}$China Innovation Center of Roche (CICoR), Shanghai 201203, China}
\date{}

\begin{document}
\maketitle
\plainfootnotetext{${}^{*}$Corresponding author: \href{mailto:zhaoheng1990@gmail.com}{zhaoheng1990@gmail.com}. ORCID: \href{https://orcid.org/0009-0004-2395-9393}{0009-0004-2395-9393}.}
\begin{abstract}
Mechanistic studies of large language models localize function in parameters, attention heads or neurons, but those parts do not specify how one inference unfolds. We treat the ordered layer-state trajectory as the empirical process and test whether layer order defines a measurable computation. Across representative Qwen, Llama and Gemma checkpoints, realized order produced paths 1.60--2.24 times as aligned with their endpoint chord as endpoint-preserving shuffles. The order effect transferred to a lexical task, but was stronger in three random initializations of the same Qwen architecture, identifying an architecture-compatible scaffold rather than a training-specific law. Direct hidden states provided stronger mechanism coordinates than vocabulary readbacks, and a task-defined coordinate made candidate competition computable. After additive interventions failed, local transition actuators selectively controlled internal trajectories across checkpoints. A confidence gate then drove 17 of 87 held-out Qwen closure trajectories across the full-vocabulary top-1 candidate boundary while leaving all 203 non-closure top-1 outputs unchanged. Safety-matched action reassignment performed similarly, locating the gain in selective gating and the bounded actuator family rather than fine-grained dose assignment. Layer order exposes an observable, computable and behaviourally consequential process of transformer inference without establishing a complete theory of LLM dynamics or control of answer correctness.
\end{abstract}
\section*{Introduction}
Large language models transform their internal state many times before producing a token. Existing work has localized components, decoded intermediate outputs and identified task-specific structures5-12,18,19, but these observations have not yet converged on a common dynamical description of inference. Model scale is one reason. Another may be more basic: an explanation depends on what is measured and on the coordinates used to compare those measurements.

Interpretability methods expose different aspects of transformer computation. Probes test decodable information, while circuit analyses isolate local computations. Causal tracing and model editing localize factual associations. Other methods use vocabulary readbacks or sparse activation features\textsuperscript{5--12}. Their conclusions depend on probe capacity, layer selection and the chosen projection. Dimensionality and similarity metrics also affect representation comparisons\textsuperscript{6,14--17}. It remains uncertain which object can be compared across layers and checkpoints while preserving the process being measured.

Intermediate vocabulary methods address several distinct questions. The tuned lens calibrates layerwise token predictions, whereas DoLa contrasts logits from different layers to improve decoding11,20. We use the same broad access route for a different purpose. Our question is whether vocabulary-facing coordinates preserve mechanism-relevant geometry because of their learned row identities. Row permutation and matched random projection make that claim experimentally separable from the usefulness of token-level readout.

We examined whether part of this difficulty is an object-coordinate mismatch. Earlier work has decoded intermediate predictions, compared representation geometry and traced particular computations\textsuperscript{11,15--20}. We treated the choice of empirical object as part of the mechanism claim and tested it directly. The decisive question was whether a proposed coordinate retained its explanatory role when its learned identity was removed. This turns coordinate choice from an analytical convention into a falsifiable hypothesis.

The experimental main line was developed in Qwen2.5-1.5B-Instruct. Gemma-2-2B-it and Llama-3.2-1B-Instruct were then used for cross-checkpoint replication and boundary tests. The three checkpoints differ substantially in depth, hidden width and vocabulary size; cross-model panels should therefore be read as tests of recurrence under architectural variation, not as three independent discovery sequences.

\begin{CodeBlock}

Study checkpoints
Main line:   Qwen/Qwen2.5-1.5B-Instruct
             1.5B | 28 layers | d=1536 | vocab=151,936 | context=32,768*
Cross-model: google/gemma-2-2b-it
             2B | 26 layers | d=2304 | vocab=256,000 | context=8,192
Cross-model: meta-llama/Llama-3.2-1B-Instruct
             1B | 16 layers | d=2048 | vocab=128,256 | context=131,072
*Qwen tokenizer model_max_length=131,072; model max_position_embeddings=32,768.
\end{CodeBlock}

The study began with TopK vocabulary readbacks because they made intermediate states inspectable at every layer. Their changes across depth exposed a reproducible late transition and raised questions about recurrent transformations and functional windows. We then tested the apparent privilege of the readback. If learned vocabulary identity supplied the relevant geometry, changing row identity should destroy it. Row permutation left preservation nearly unchanged, while matched random projections often preserved as much or more. This result rejected the learned-row explanation and shifted the empirical object to the hidden states from which the readbacks were derived. Grouped, leakage-controlled and capacity-matched comparisons established ordered high-dimensional layer states as stronger coordinates for controlled mechanism-label organization.

We then asked what dynamical mechanism those trajectories revealed. Across all three checkpoints, candidate-relative support evolved within functional windows, and real layer order was more directionally aligned and less curved or detoured than order-shuffled controls. Large residuals ruled out exact-geodesic motion but carried complementary mechanism information. Trajectory-informed actions produced target-specific changes in an internal dominance endpoint beyond fixed and matched shuffled policies. Together, these experiments reveal inference as a high-dimensional, geodesic-biased competitive flow with structured off-direction processing. The picture is reproducible across the tested checkpoints even though its effect sizes are checkpoint dependent. It provides a mechanism-level account of internal inference dynamics rather than a complete dynamical law.

The study makes four testable contributions. It falsifies learned vocabulary-row identity as the source of the preserved geometry; establishes the ordered layer-state trajectory as the primary mechanism object; characterizes inference as geodesic-biased competitive flow with structured residual motion; and demonstrates that trajectory-informed policies discriminate matched from mismatched internal actions. Each contribution is tied to a projection, capacity, order or intervention control.

\section*{Core concepts}
The mechanism picture uses a small set of coordinates. Their roles are fixed throughout the paper.

\begin{table}[H]
\centering\scriptsize\sloppy
\begin{tabularx}{\textwidth}{@{}XX@{}}
\toprule
Term & Operational meaning \\
\midrule
Layer-state trajectory & The ordered sequence of hidden states across transformer layers for one checkpoint and prompt. “Hidden-state trajectory” denotes the same object; “layer-state trajectory” is the canonical term used in claims. \\
Functional window & A checkpoint-specific layer interval selected by a declared functional objective rather than by nominal layer identity. \\
Candidate trajectories & The leading and competing candidate-relative paths declared by the controlled prompt construction. \\
\(\Delta U\) & A signed low-rank scalar measuring relative support for the leading versus competing candidate trajectory within the decision window. \\
Observed-transition coordinate \(O_l^{\mathrm{cont}}\) & A continuous latent coordinate derived from the realized transition signature from layer \(l\) to \(l+1\); it describes rather than predicts that transition. \\
Target-excluded estimate \(\widehat O_l^{\mathrm{cont}}\) & An estimate of the latent transition coordinate constructed without target-layer information; it tests advance recoverability. \\
Geodesic-biased flow & Greater alignment and lower curvature or detour than the declared order-shuffle control, without exact shortest-path motion. \\
Structured residual & Off-reference-direction motion that retains stable candidate or mechanism information beyond the aligned component. \\
\bottomrule
\end{tabularx}
\end{table}

\section*{Results overview}
This study reveals a mechanism-level picture of internal transformer inference. Candidate trajectories compete within a high-dimensional layer-state space, and their relative support evolves through checkpoint-specific functional windows. The flow has a reproducible global organization: real layer order is more aligned and less curved or detoured than order shuffles. It is not an exact geodesic. Structured off-direction motion retains complementary mechanism information, and trajectory-informed actions produce target-specific internal changes beyond fixed and matched shuffled policies. Learned vocabulary-row identity fails as the source of this organization; the ordered hidden-state trajectory is the primary mechanism object. The same dynamical form recurs across three representative checkpoints, while its magnitude and the relative strength of particular coordinates remain checkpoint dependent.

The graphical overview gives a low-dimensional orientation aid. The surface, paths and competition ridge are illustrative; the evidence comes from high-dimensional layer-state arrays, projection controls, grouped comparisons, route-order nulls and held-out operational tests. The Results then follow the discovery sequence. Vocabulary readbacks first supplied an observational window, operator and depth controls narrowed that interpretation, and projection controls rejected learned vocabulary-row identity as the source of the preserved geometry. That failed explanation exposed the ordered hidden-state trajectory as the more fundamental empirical object.

\begin{figure}[H]
\centering
\includegraphics[width=\textwidth]{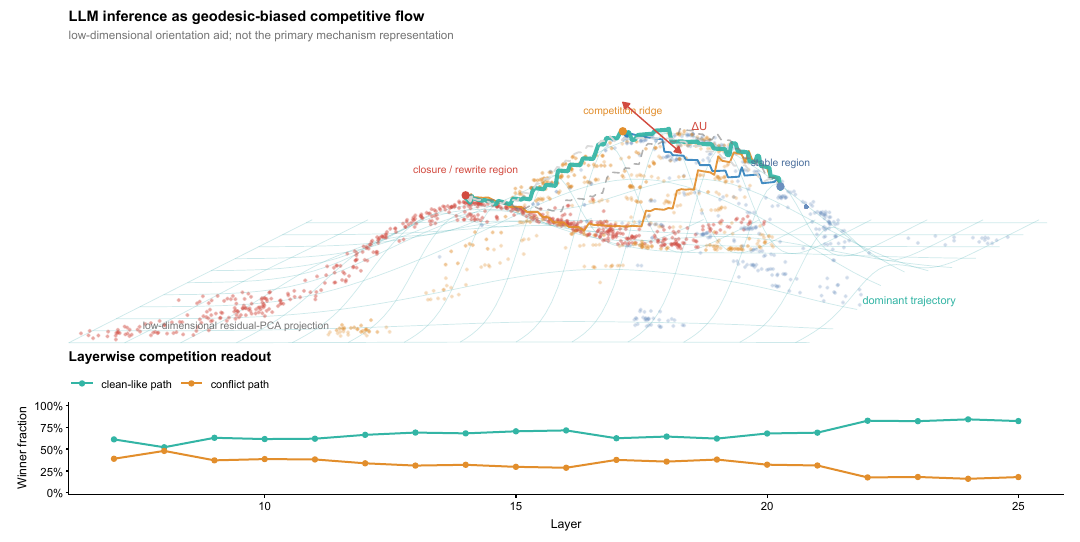}
\caption*{\textbf{Graphical overview | Candidate-trajectory competition during transformer inference.} This low-dimensional residual-PCA display is an orientation aid for the retained conclusion. It illustrates layerwise competition, the bounded dominance observable $\Delta U$, and the distinction between dominant and competing paths. The surface and paths are not literal estimates of a semantic manifold, and the mechanism claim does not rest on this projection. Subsequent figures test the account in declared high-dimensional layer-state coordinates and under projection, order-shuffle, leakage and intervention controls.}
\end{figure}
\FloatBarrier
\section*{Quick guide to the evidence chain}
\begin{table}[H]
\centering\scriptsize\sloppy
\begin{tabularx}{\textwidth}{@{}XXX@{}}
\toprule
Figures & Question answered & Retained result \\
\midrule
Figs. 1-4 & What can vocabulary-facing readbacks reveal across layers and checkpoints? & A reproducible observational flow and coarse late transition motivate functional alignment, while fine operator labels largely track depth. \\
Figs. 5-7 & Can candidate competition be compressed and detected before the decision window? & \(\Delta U\) captures candidate-relative dominance, with leakage-reduced precursor organization that varies by checkpoint. \\
Fig. 8 & Does learned vocabulary-row identity supply the observed geometry? & Row permutation and matched projection controls reject learned row identity as the source. \\
Figs. 9-11 & Which empirical object best retains the mechanism organization? & Grouped and matched-capacity audits upgrade the primary object to the ordered high-dimensional layer-state trajectory. \\
Figs. 12-14 & What dynamical organization does the trajectory exhibit? & Descriptive transition closure, geodesic-biased flow and structured non-geodesic residual motion define the retained mechanism picture. \\
Fig. 15 & Is the trajectory organization operationally discriminating? & Trajectory-informed policies produce target-specific internal modulation beyond fixed and matched shuffled alternatives. \\
\bottomrule
\end{tabularx}
\end{table}

\section{Opening a layer-resolved observational window}
\subsubsection*{Could vocabulary-facing readbacks make hidden computation observable?}
The first obstacle was observational. Intermediate activations were available, but there was no agreed coordinate for inspecting their evolution without first choosing a circuit, label or task-specific probe. The vocabulary projection attached to each checkpoint offered a possible window. If a TopK readback retained organized information, its layerwise profile should recur across seeds and remain recognizable under changes to TopK and prompt form. Projecting each hidden state through the output projection produced a layer-indexed vocabulary neighbourhood that made this premise testable (Fig. 1a). It was our first candidate coordinate, not yet a causal object.

\begin{figure}[!htbp]
\centering
\includegraphics[width=0.98\textwidth,height=0.76\textheight,keepaspectratio]{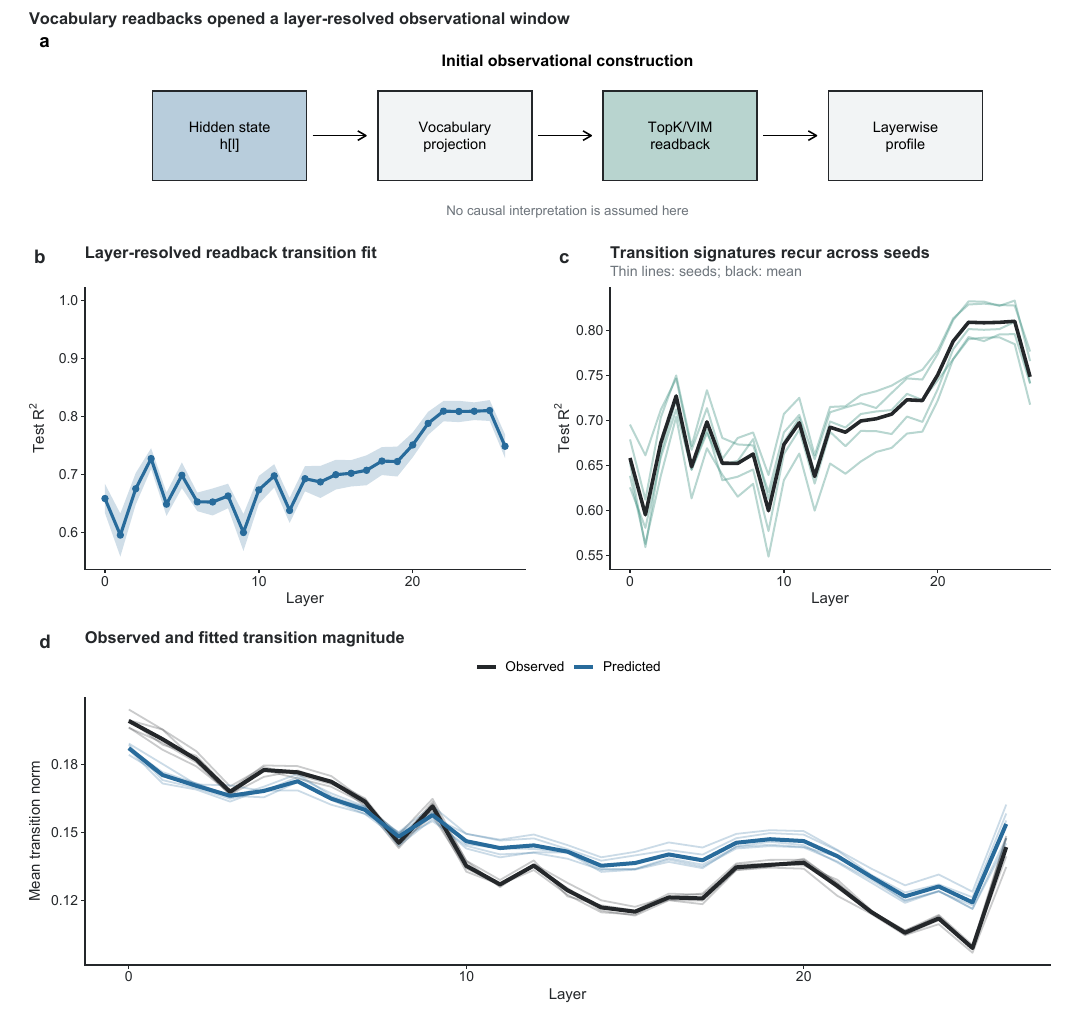}
\caption{Vocabulary readbacks as a layer-resolved observational window. \textbf{a,} Construction schematic showing the mapping from a hidden state at layer \(l\) to a vocabulary projection, TopK/VIM readback and layerwise profile. The schematic defines an observable and does not assign causal status to the projection. \textbf{b,} Mean held-out transition-fit \(R^2\) by layer; ribbons show 95\% confidence intervals across five seeds. Each split contained 196 observations, with 147 training and 49 test observations. \textbf{c,} Layerwise transition signatures for individual seeds (thin teal lines) and their mean (black line). \textbf{d,} Observed and fitted mean transition magnitudes across layers; thin lines show individual seeds and thick lines show seed means. See the corresponding Source Data figure and the Source Data manifest.}
\label{fig:1}
\end{figure}

\FloatBarrier

The vocabulary-facing neighbourhoods changed reproducibly with layer depth. Their profiles recurred across the available seeds and survived the tested TopK and prompt perturbations (Figs. 1 and 2). Natural-language transformations preserved high centre cosine while producing larger variation in overlap and path agreement, separating stable readback geometry from condition-sensitive neighbourhood composition (Fig. 2d). Complete prompt, seed, TopK and negative-example summaries are reported in Supplementary Figs. S1-S3.

\begin{figure}[!htbp]
\centering
\includegraphics[width=0.98\textwidth,height=0.76\textheight,keepaspectratio]{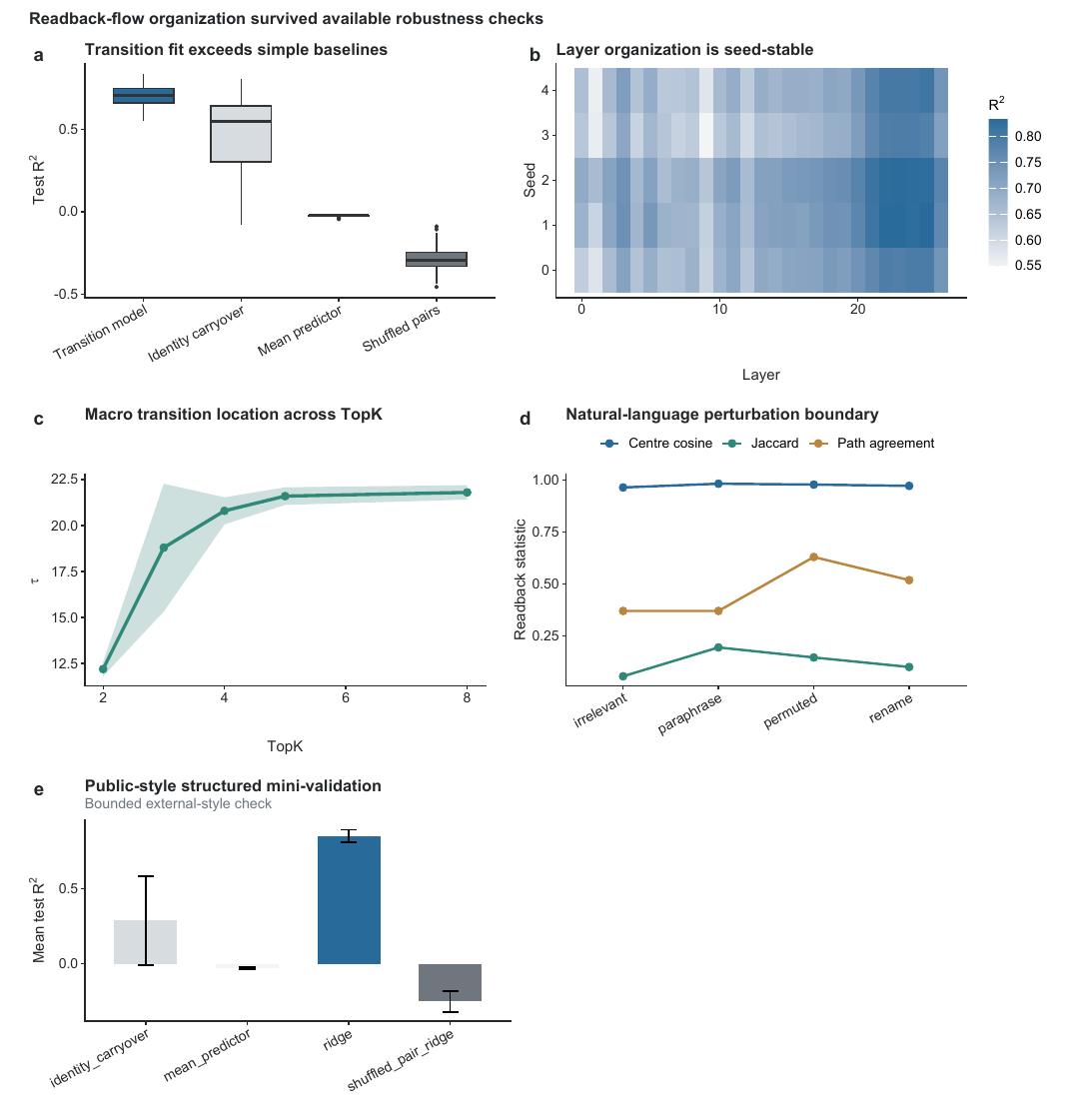}
\caption{Robustness of readback-flow organization. \textbf{a,} Held-out \(R^2\) distributions across layers and seeds for the fitted transition model and identity-carryover, mean-predictor and shuffled-pair baselines. \textbf{b,} Seed-by-layer heat map of held-out transition-model \(R^2\). \textbf{c,} Estimated late macro-transition location \(\tau\) across TopK values; ribbons show 95\% confidence intervals across seeds. \textbf{d,} Centre cosine, Jaccard overlap and path agreement under four natural-language perturbation conditions. These statistics have different definitions and should be compared across conditions within a metric, not across metric types. \textbf{e,} Mean held-out \(R^2\) in the public-style structured mini-validation; error bars show the reported standard deviation. This panel is a bounded external-style check and not a benchmark-generalization result. See the corresponding Source Data figure and the Source Data manifest.}
\label{fig:2}
\end{figure}

\FloatBarrier

\begin{figure}[!htbp]
\centering
\includegraphics[width=0.98\textwidth,height=0.76\textheight,keepaspectratio]{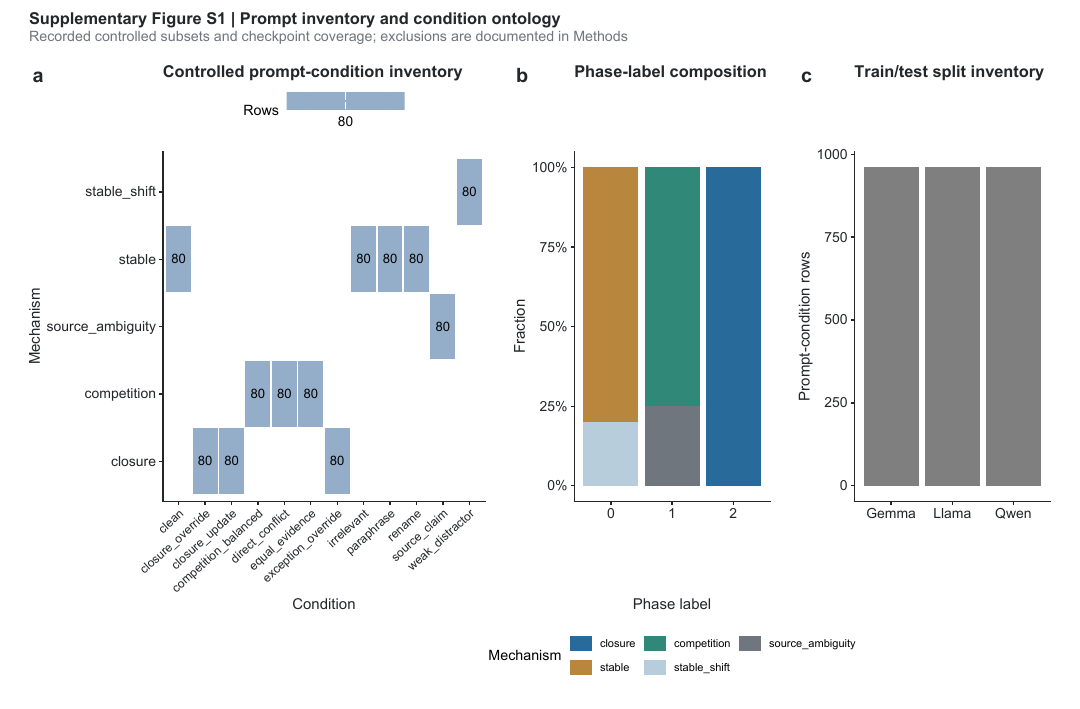}
\caption*{Prompt inventory and condition ontology. \textbf{a,} Counts in the controlled prompt-condition inventory by recorded condition and mechanism label. \textbf{b,} Mechanism composition within each phase label. \textbf{c,} Graph-grouped internal-modulation split inventory for the three representative checkpoints. Tokenization and exclusion rules are specified in Supplementary Methods. See Source Data Figs. 9 and 15 and the Source Data manifest.}
\end{figure}

\FloatBarrier

\begin{figure}[!htbp]
\centering
\includegraphics[width=0.98\textwidth,height=0.76\textheight,keepaspectratio]{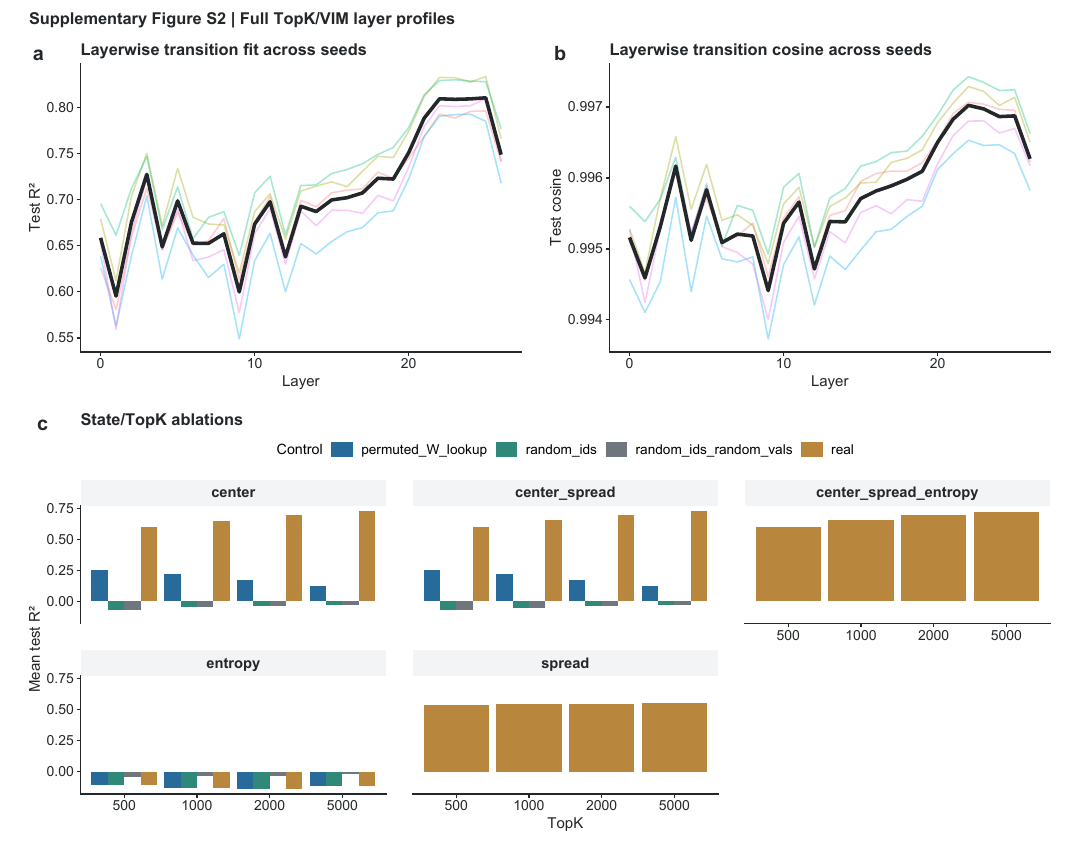}
\caption*{Full TopK/VIM layer profiles. \textbf{a,b,} Layerwise transition-fit R2 and cosine values across recorded seeds; black curves show seed means. \textbf{c,} State/TopK ablations across feature modes and TopK values. Readbacks remain diagnostics rather than causal mechanism variables. See Source Data Figs. 1-3 and the Source Data manifest.}
\end{figure}

\FloatBarrier

Vocabulary-facing readbacks thus provided a reproducible, layer-resolved observational surface. Static similarity, however, could not distinguish neighbourhoods reached through different transformations. Having the same readout at every layer made those transformations measurable and raised a new question: did successive layers reuse a limited organization? At this point, it was still plausible that the learned output rows exposed structure used during inference.

\subsubsection*{Did adjacent readback changes reuse a limited repertoire of transformations?}
Reproducible readbacks showed which vocabulary entries were visible at each layer. The rearrangement between layers remained unspecified. We represented each readback by its declared centre and spread, then fitted adjacent-layer maps \(\hat F_l^k:Z_l^k\rightarrow Z_{l+1}^k\). A measurable flow should retain positive held-out fit outside the fitting sample. The maps met this criterion across the stack (Figs. 1b-d and 3b).

The fitted maps suggested a stronger possibility: inference might reuse a small repertoire of layer-to-layer transformations. We clustered their signatures across depth to test this idea. The first path resembled three broad regimes, provisionally labelled input boundary, bulk and output boundary (Fig. 3a). In Qwen, persistent entry into the final regime began at L21-to-L22, pointing to a late macro transition.

\begin{figure}[!htbp]
\centering
\includegraphics[width=0.98\textwidth,height=0.76\textheight,keepaspectratio]{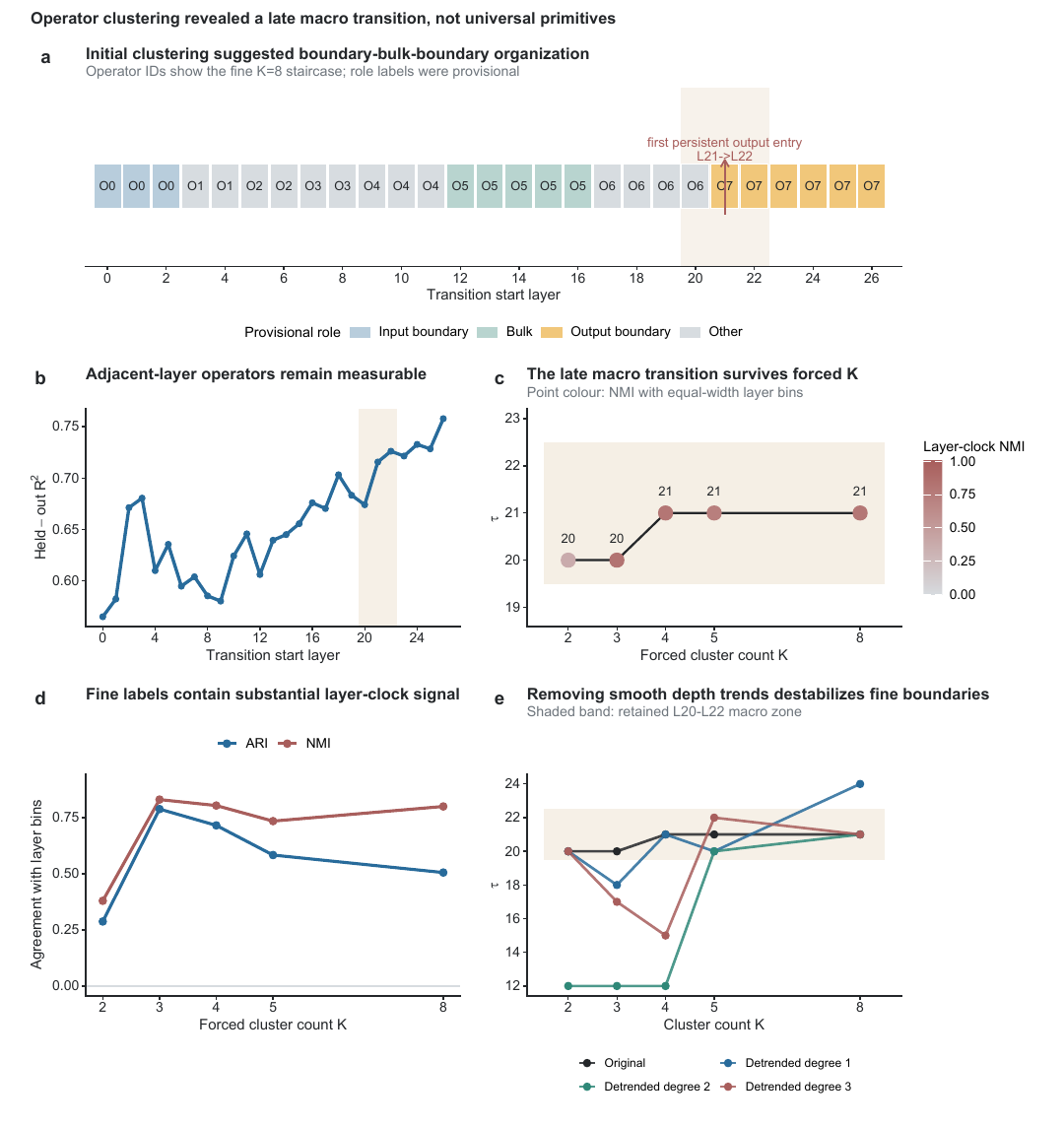}
\caption{Operator clustering revealed a late macro transition, not universal primitives. \textbf{a,} Initial K=8 clustering path across the 27 adjacent-layer transitions of the Qwen checkpoint. Tiles show operator labels O0-O7; colours show the provisional role assignment used at this discovery stage. The first persistent entry into the output-regime cluster occurred at L21-to-L22. Role assignments were exploratory and are not interpreted as primitive causal operators. \textbf{b,} Held-out \(R^2\) of adjacent-layer VIM operator fits across depth. The pale band marks transition start layers 20-22. \textbf{c,} Estimated late transition \(\tau\) under forced cluster counts \(K=2,3,4,5,8\). The shaded interval marks the retained L20-L22 macro zone; point colour reports normalized mutual information (NMI) between operator labels and equal-width layer bins. \textbf{d,} Adjusted Rand index (ARI) and NMI between the original operator labels and equal-width layer bins. High agreement at several cluster resolutions indicates substantial layer-clock structure in the fine labels. \textbf{e,} Estimated \(\tau\) before and after polynomial removal of smooth layer trends. Original estimates remain within L20-L21 across forced K, whereas detrended fine boundaries vary widely. Together, the controls narrow the initial operator-family interpretation to organized readback flow with a robust late macro transition; they do not support a universal catalogue of primitive operators. The analysis used 196 prompt-condition instances, with 147 training and 49 held-out instances for the adjacent-layer fits. Operator clustering used the declared Audit-6A.1 construction. The forced-K values, ARI, NMI and detrended boundary estimates are shown without inferential p-values; they are sensitivity and confound diagnostics. See Source Data Fig. 3 and the Source Data manifest.}
\label{fig:3}
\end{figure}

\FloatBarrier

Smooth change with depth can also produce discrete-looking clusters. We tested this alternative with forced resolutions \(K=2,3,4,5,8\), equal-width layer bins, layer-index prediction and depth-detrended reclustering. The fine labels contained substantial layer-clock information (Fig. 3c-e), restricting the three-regime interpretation to an exploratory stage model. A coarser feature remained: the transition estimates were L20, L20, L21, L21 and L21 across the five resolutions (Fig. 3c).

The transition operators remained measurable throughout depth, with held-out \(R^2\) varying systematically and rising in later layers (Fig. 3b). Additional null-segmentation, state/TopK-ablation and relation-preserving analyses showed that the retained macro organization was not reproduced by the tested segmentations and depended on both selected rows and readback features (Supplementary Fig. S3). These checks further separated near-invariant centre geometry from more variable neighbourhood overlap.

\begin{figure}[!htbp]
\centering
\includegraphics[width=0.98\textwidth,height=0.76\textheight,keepaspectratio]{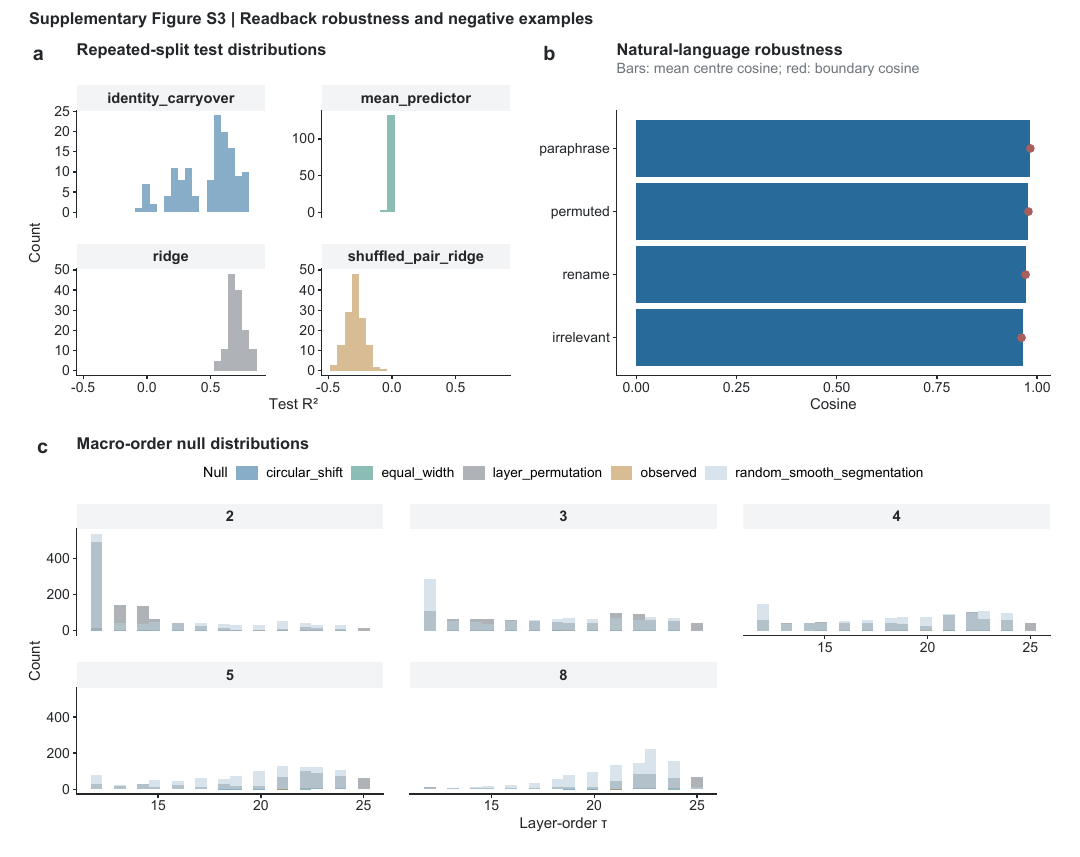}
\caption*{Readback robustness and negative examples. \textbf{a,} Repeated-split test distributions by baseline. \textbf{b,} Natural-language perturbation results. \textbf{c,} Macro-order null distributions across K. Weak and negative controls are retained rather than removed. See Source Data Figs. 2-3 and the Source Data manifest.}
\end{figure}

\FloatBarrier

The layer-clock result changed our interpretation while leaving the observation intact. Readbacks participated in a measurable layerwise flow, and a late macro transition survived. Evidence for a small catalogue of primitive operators was absent. The strongest transitions also occurred at different nominal depths in different checkpoints. Recurrence could no longer be defined by a detailed cluster label or a common layer number, which shifted the comparison to functional windows.

\subsubsection*{How can the same functional question be compared across checkpoints of different depth?}
Cross-checkpoint comparison created a coordinate problem. Nominal layer matching would assume the correspondence under test. Normalized depth would instead assume uniform scaling of function with model depth. We therefore scanned all eligible intervals within each checkpoint and selected windows by the same declared functional objective, without requiring common layer indices or widths. Complete scans showed that each checkpoint and objective selected a different interval (Fig. 4a-c and Supplementary Fig. S4). We call these checkpoint-specific intervals functional windows. Subsequent comparisons used these windows. Selection and evaluation used the same controlled dataset, so best-window values remain descriptive summaries rather than independently confirmed optima. This is functional-role alignment, not evidence for representational homology or a conserved invariant.

\begin{figure}[!htbp]
\centering
\includegraphics[width=0.98\textwidth,height=0.76\textheight,keepaspectratio]{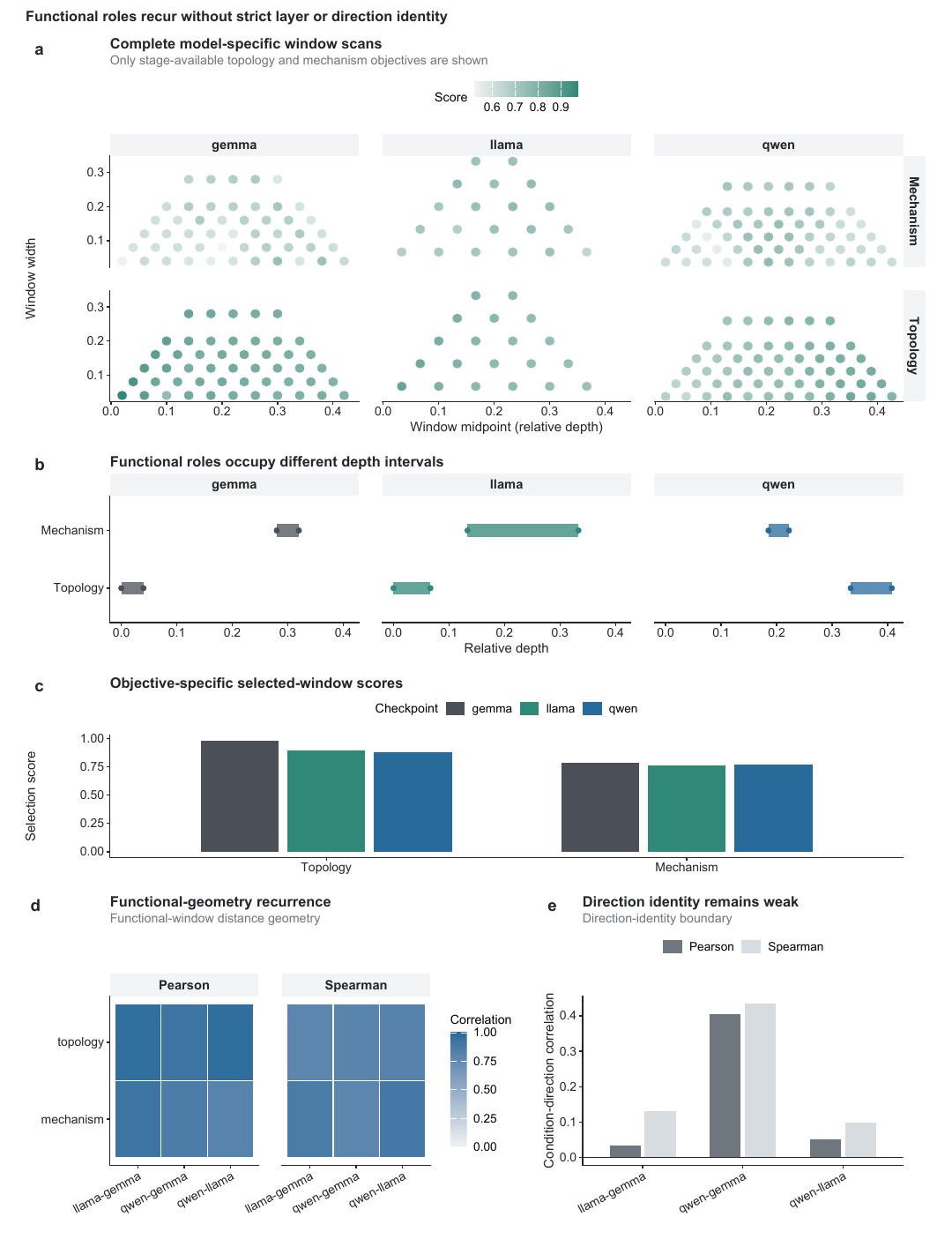}
\caption{Functional recurrence without strict layer or direction identity. \textbf{a,} Complete model-specific window scans for topology and mechanism objectives in representative Gemma, Llama and Qwen checkpoints. Colour denotes the corresponding selection score. Downstream dominance fields are not used because they become available only after \(\Delta U\) is introduced later in the discovery chain. \textbf{b,} Selected topology and mechanism intervals in relative layer depth. \textbf{c,} Objective-specific scores of the selected intervals. \textbf{d,} Pairwise recurrence of functional-window distance geometry measured by Pearson and Spearman correlations. \textbf{e,} Condition-direction correlations from the separate direction-identity audit. Functional geometry recurred strongly, whereas strict direction identity was weak and heterogeneous. See the corresponding Source Data figure and the Source Data manifest.}
\label{fig:4}
\end{figure}

\FloatBarrier

\begin{figure}[!htbp]
\centering
\includegraphics[width=0.98\textwidth,height=0.76\textheight,keepaspectratio]{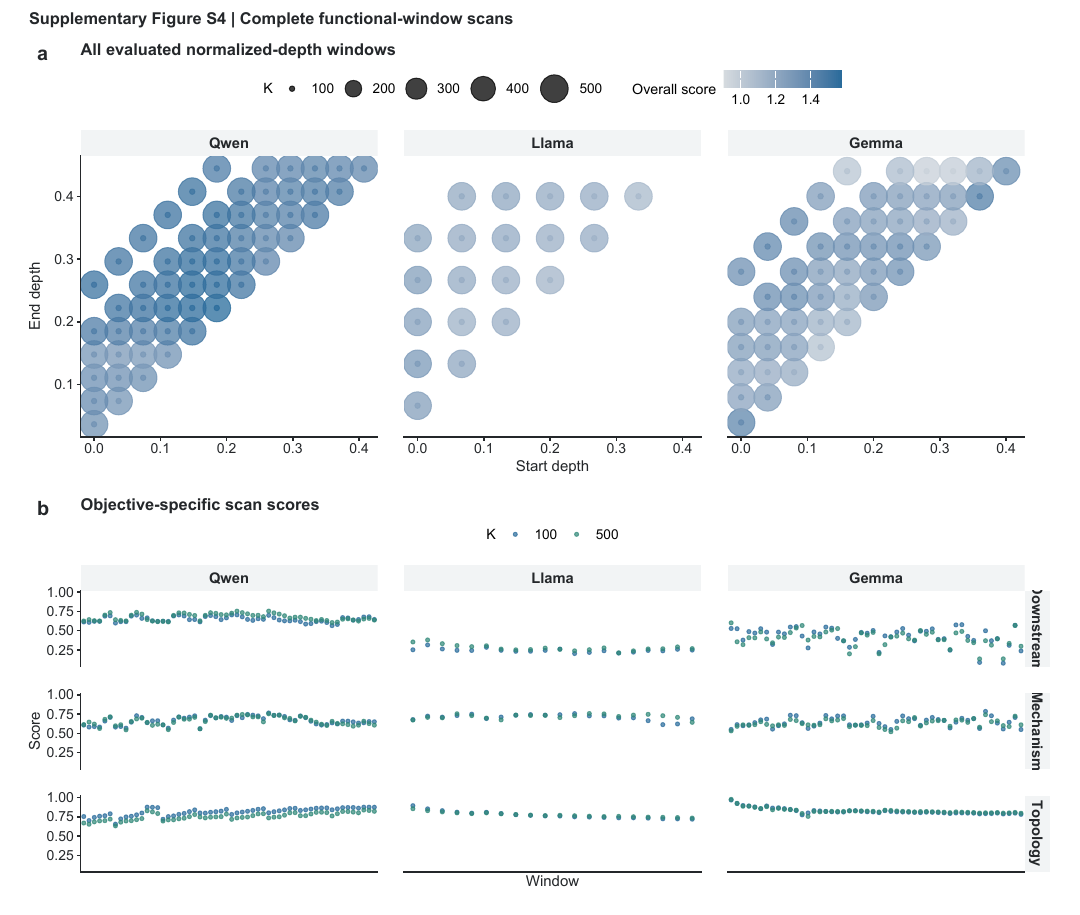}
\caption*{Complete functional-window scans. \textbf{a,} All evaluated windows in normalized checkpoint depth, with colour denoting overall score and point size denoting K. \textbf{b,} Objective-specific scores for topology, mechanism-label separation and downstream association. Window recurrence does not imply layer identity. See Source Data Fig. 4 and the Source Data manifest.}
\end{figure}

\FloatBarrier

Objective-specific window geometry recurred despite this heterogeneity (Fig. 4d). Direction identity was weak. Pairwise correlations of condition-direction distances were weak or checkpoint dependent, whereas functional-geometry correlations were stronger (Fig. 4e and Supplementary Fig. S5). We therefore compared functional roles without assuming common coordinates. The selected windows contained many correlated variables, and their relation to the earlier operator reuse suggested that a continuous accumulation process might admit a compact candidate-relative coordinate.

\begin{figure}[!htbp]
\centering
\includegraphics[width=0.98\textwidth,height=0.76\textheight,keepaspectratio]{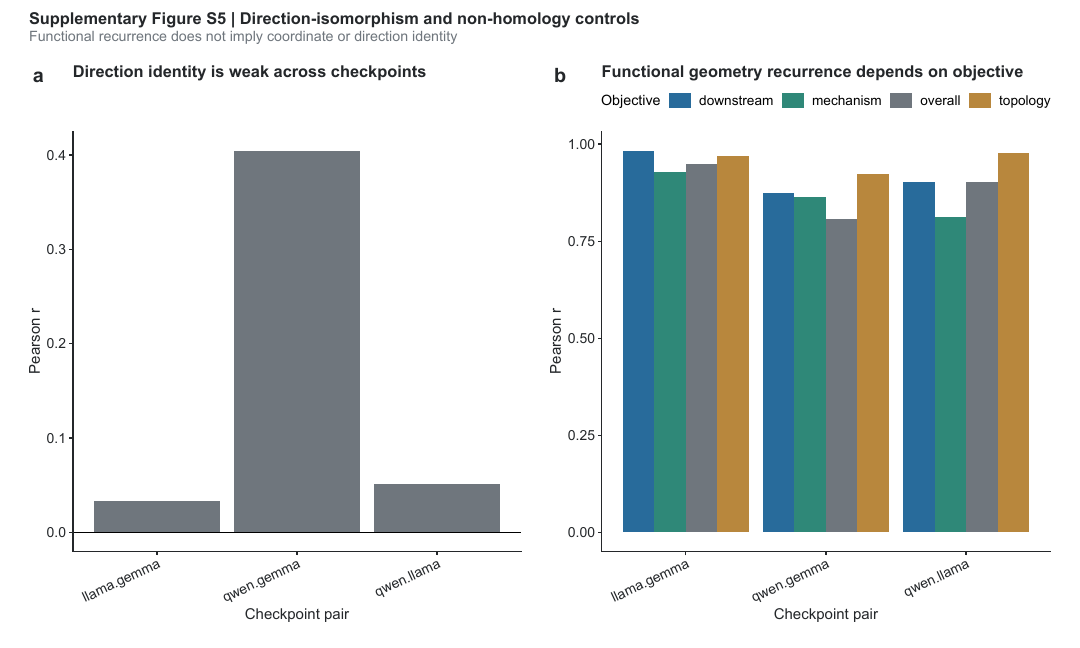}
\caption*{Direction-isomorphism and non-homology controls. \textbf{a,} Cross-checkpoint correlations of condition-direction distances. \textbf{b,} Recurrence of functional geometry by selection objective. These controls support functional recurrence but reject coordinate or direction identity. See Source Data Fig. 4 and the Source Data manifest.}
\end{figure}

\FloatBarrier

\section{Challenging the readback account}
\subsubsection*{Can many local flow measurements be reduced to one bounded dominance question?}
Functional alignment left many local flow measurements to compare. If they reflected a shared candidate-competition process, their clean-relative decision matrix should have a dominant low-rank direction. The first component accounted for 92\% of variance in Qwen, 83\% in Llama and 97\% in Gemma (Fig. 5a,d). Its loadings extended across a continuous late window (Fig. 5b). The oriented scores separated the negative, critical and positive response regimes (Fig. 5c).

\begin{figure}[!htbp]
\centering
\includegraphics[width=0.98\textwidth,height=0.76\textheight,keepaspectratio]{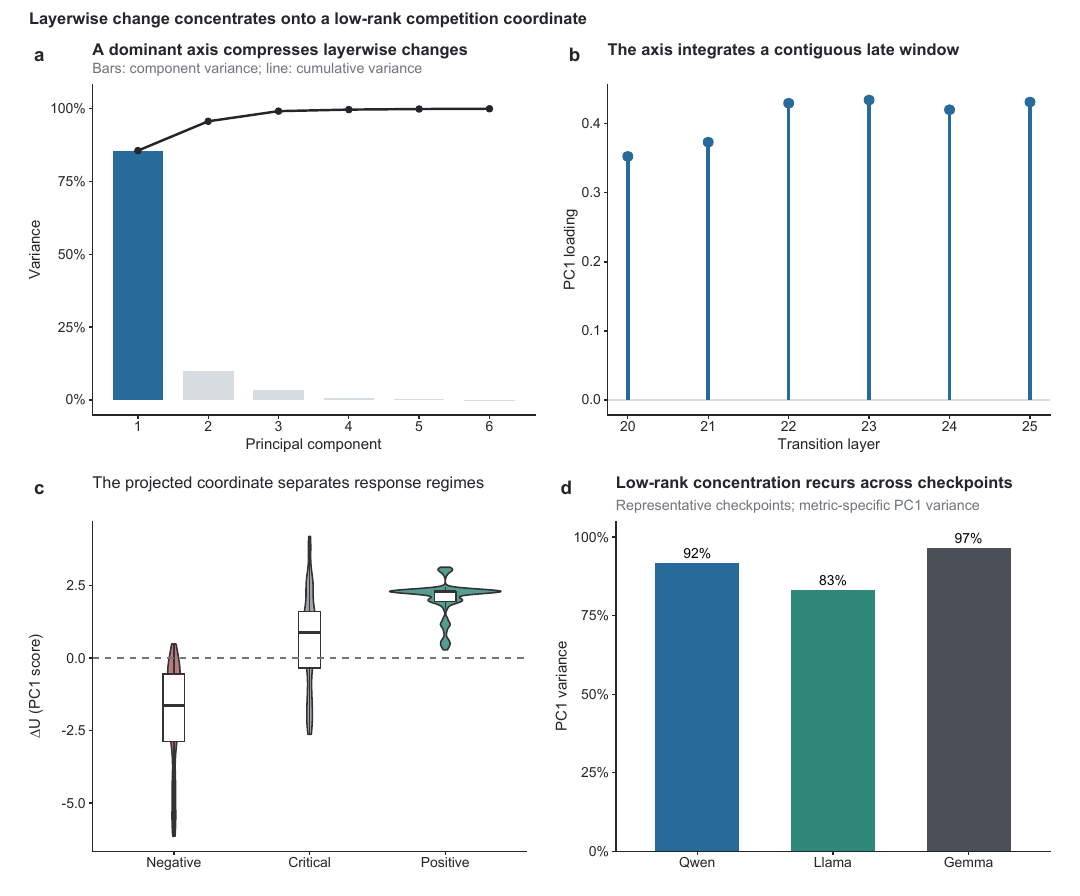}
\caption{Layerwise change concentrates onto a low-rank competition coordinate. \textbf{a,} PCA spectrum of six late-window layer-to-layer readback changes in the original low-rank dominance analysis. Bars show variance explained by each component; the line shows cumulative variance. \textbf{b,} Loadings of transitions L20-L25 on PC1. \textbf{c,} Distribution of the oriented PC1 coordinate, subsequently denoted \(\Delta U\), across negative, critical and positive response regimes (\(n=768\) prompt-condition observations). Violin shapes show empirical density and boxes show medians and interquartile ranges. \textbf{d,} Fraction of variance concentrated on PC1 in representative Qwen, Llama and Gemma checkpoints. Cross-checkpoint values are metric-specific and do not imply a shared physical axis. See Source Data Fig. 5 and Source Data manifest.}
\label{fig:5}
\end{figure}

\FloatBarrier

We defined the signed candidate-path separation as \(\Delta U\), a low-rank coordinate that summarizes relative support for the leading and competing candidate trajectories within a declared window. It is a bounded dominance observable. Let \(D\) denote the clean-relative decision matrix over the permitted decision-window features. We oriented its leading principal direction towards greater support for the leading trajectory. The competing trajectory supplied the reference. Final answer labels were excluded as an independent validation target. Supplementary Fig. S6 reports the loadings, sign audit and alternative scaling.

\begin{figure}[!htbp]
\centering
\includegraphics[width=0.98\textwidth,height=0.76\textheight,keepaspectratio]{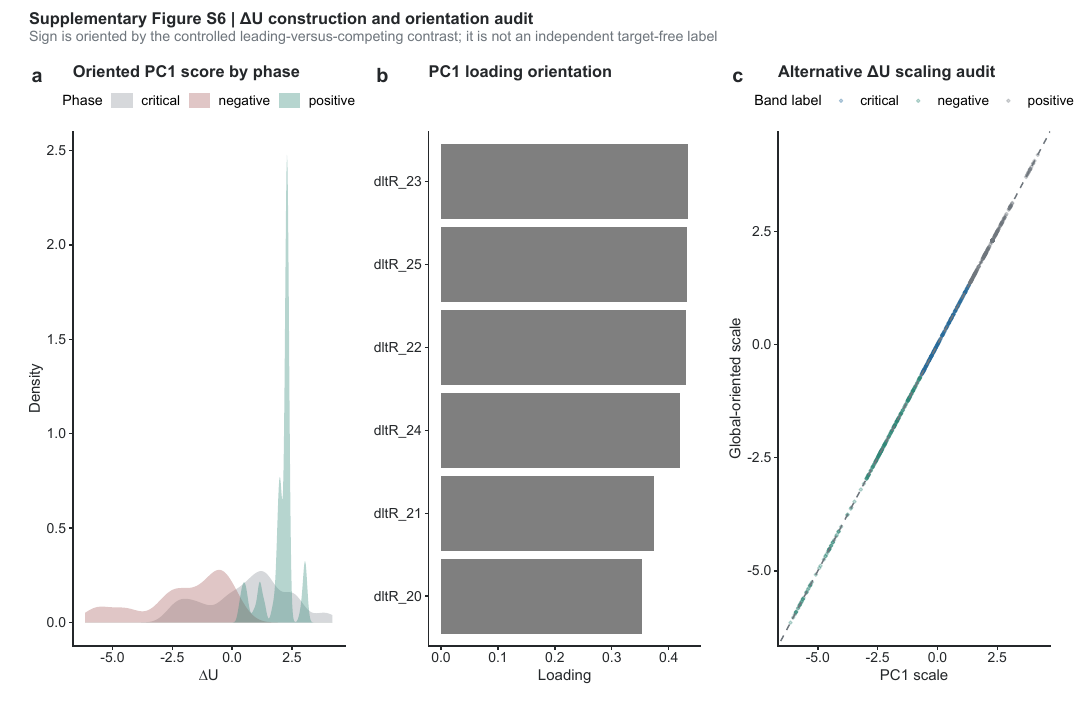}
\caption*{\(\Delta U\) construction and orientation audit. \textbf{a,} Oriented PC1 score distributions by controlled phase. \textbf{b,} PC1 loadings on the decision-window contrast features. \textbf{c,} PC1 and global-oriented scalings. The sign of \(\Delta U\) is oriented by the controlled leading-versus-competing contrast; \(\Delta U\) remains a bounded observable and is not an independently target-free label. See Source Data Figs. 5-6 and the Source Data manifest.}
\end{figure}

\FloatBarrier

The \(\Delta U\) distributions formed checkpoint-specific competition bands. Negative, critical and positive conditions overlapped in the full feature space but were ordered along the oriented coordinate (Fig. 6a,b). Band occupancy varied by condition, and a band-only classifier reached macro-F1 0.61 (Fig. 6c,d). Because the coordinate and condition contrast shared the controlled candidate-pair design, this agreement was partly expected. We treated it as a construction check, not independent evidence for a mechanism, correctness variable or universal order parameter. The more demanding test was whether related structure appeared before the decision window used to define \(\Delta U\).

\begin{figure}[!htbp]
\centering
\includegraphics[width=0.98\textwidth,height=0.76\textheight,keepaspectratio]{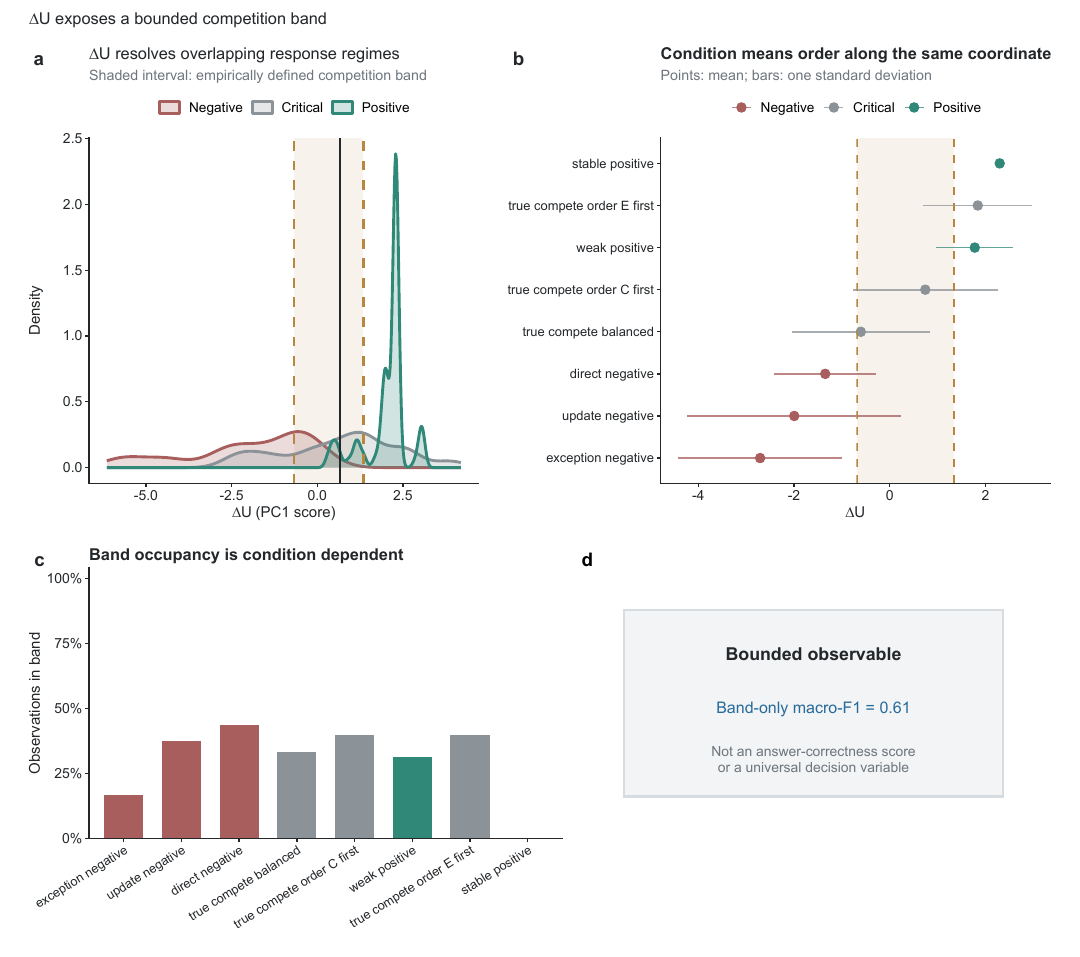}
\caption{\(\Delta U\) exposes a bounded competition band. \textbf{a,} Empirical \(\Delta U\) distributions for negative, critical and positive response regimes. The shaded interval is the empirically defined competition band \([U_-,U_+]\); the solid line marks its critical centre. \textbf{b,} Condition-level means and one standard deviation, ordered along \(\Delta U\). \textbf{c,} Fraction of observations from each condition falling inside the band. \textbf{d,} Boundary summary. A classifier using the band midpoint alone achieved macro-F1 = 0.61 in this controlled dataset. \(\Delta U\) is a bounded observable, not an answer-correctness score or a universal decision variable. See Source Data Fig. 6 and Source Data manifest.}
\label{fig:6}
\end{figure}

\FloatBarrier

\subsubsection*{Does dominance-relevant structure exist before the decision window?}
The compactness of \(\Delta U\) created an immediate circularity risk. A predictor using the decision window could appear informative simply because it reread the features used to construct the target. We therefore set a threat-driven criterion: a precursor interpretation required some association to survive both temporal exclusion of the decision window and removal of R-derived candidate-pair features. We first retained only predecision transitions, then restricted the predictors to non-R vocabulary-neighbourhood features. All rows sharing a relation graph were kept in the same split, and uncertainty was estimated by resampling relation-graph groups with model refitting.

Association with later \(\Delta U\) remained after the predecision restriction in all three checkpoints, although its magnitude differed (Fig. 7a). Non-R features also retained positive association and mechanism-label separation (Fig. 7b). Grouped bootstraps gave strong Qwen and Gemma correlations and a weaker, less certain Llama result (Fig. 7c,d and Supplementary Fig. S7). These controls reduced direct leakage; causal status for the vocabulary readback remained unsupported. They left a more basic question unresolved: did the preserved geometry depend on learned vocabulary-row identity?

\begin{figure}[!htbp]
\centering
\includegraphics[width=0.98\textwidth,height=0.76\textheight,keepaspectratio]{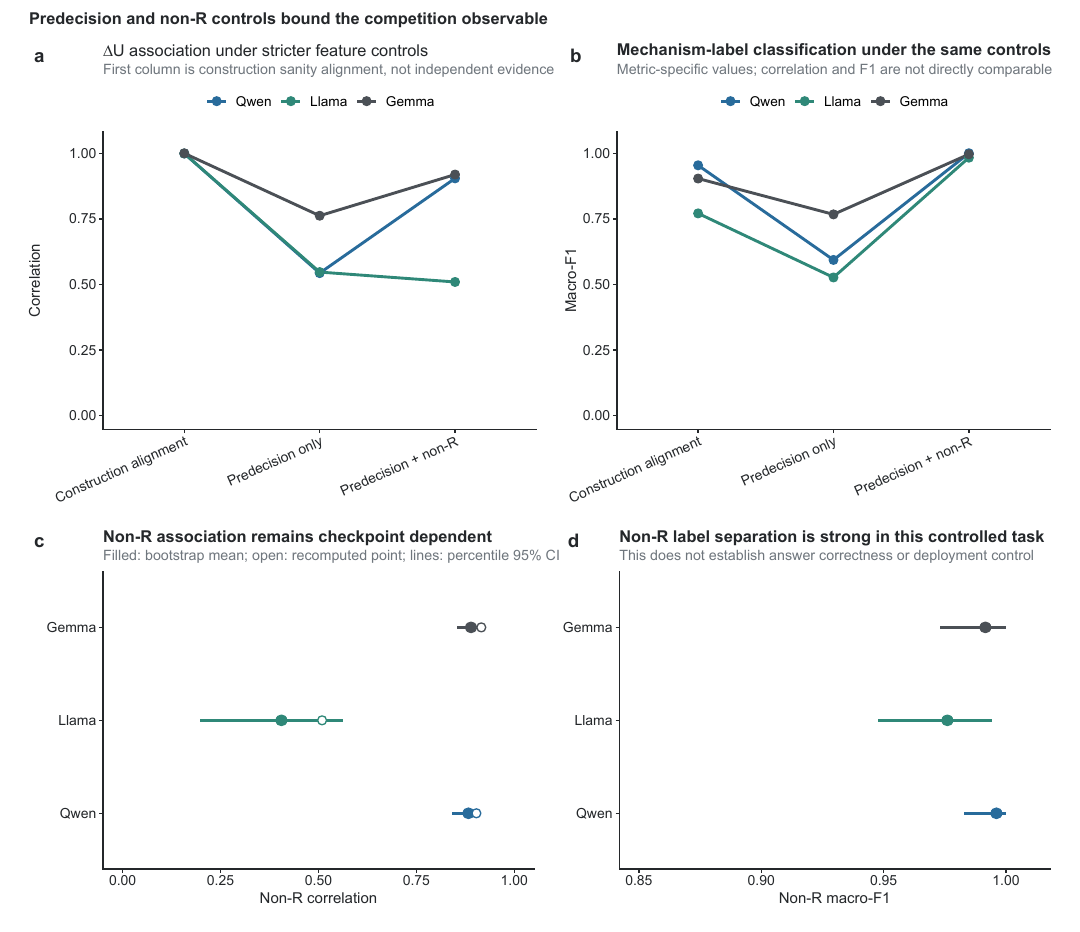}
\caption{Predecision and non-R controls bound the competition observable. \textbf{a,} Association with \(\Delta U\) across three feature constructions: unrestricted construction alignment, features restricted to transitions before the decision window, and predecision vocabulary-neighbourhood-flow features excluding R. The unrestricted value is a construction sanity alignment and is not independent mechanistic evidence. \textbf{b,} Mechanism-label macro-F1 under the same control ladder. Correlation and macro-F1 are metric-specific and are not directly comparable. \textbf{c,} Non-R correlation for the best \(\Delta U\) window. Filled points are grouped-bootstrap means, open points are recomputed full-sample estimates, and lines are percentile 95\% confidence intervals (200 successful resamples per checkpoint). \textbf{d,} Non-R mechanism-label macro-F1 for the best mechanism window, shown as bootstrap means and percentile 95\% intervals. Strong controlled-task label separation does not establish answer correctness or deployment control. See Source Data Fig. 7 and Source Data manifest.}
\label{fig:7}
\end{figure}

\FloatBarrier

\begin{figure}[!htbp]
\centering
\includegraphics[width=0.98\textwidth,height=0.76\textheight,keepaspectratio]{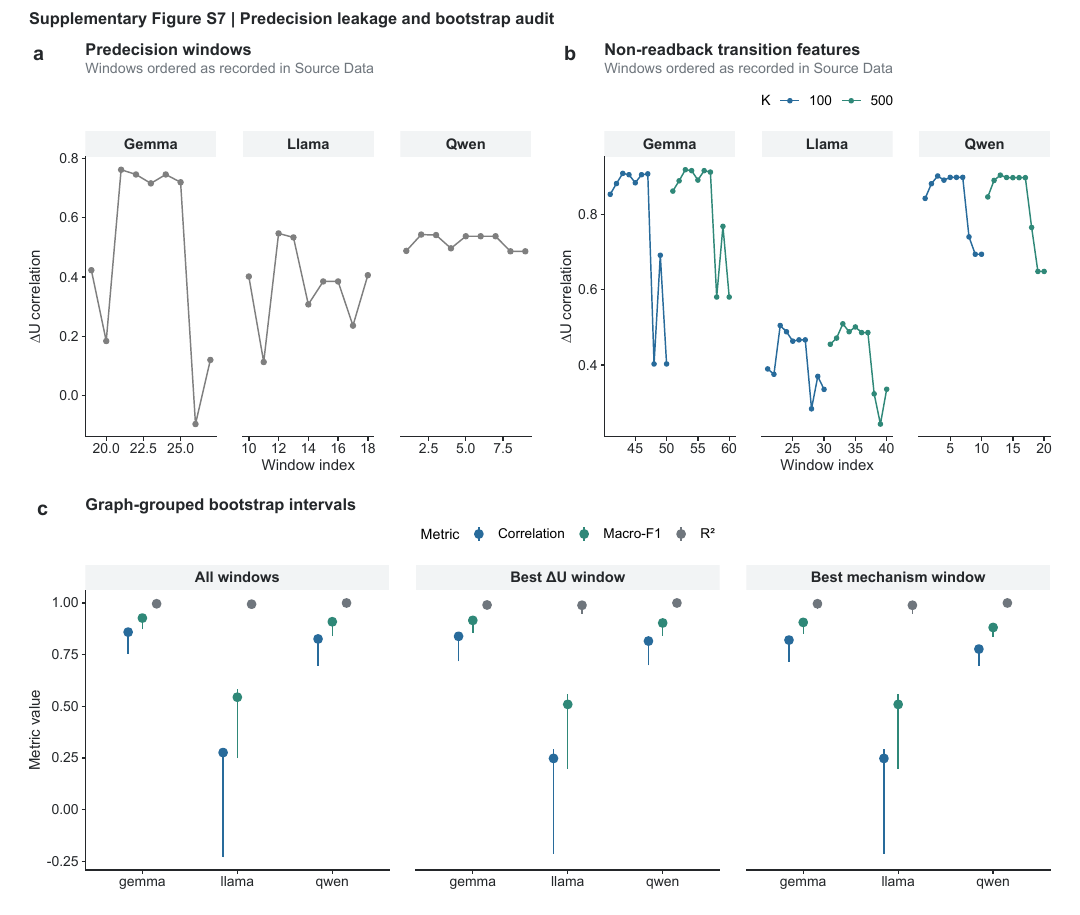}
\caption*{Predecision leakage and bootstrap audit. \textbf{a,} Predecision hidden-flow windows, all excluding the decision window. \textbf{b,} Non-readback neighbourhood-transition windows. Window indices preserve Source Data order. \textbf{c,} Graph-grouped bootstrap intervals for correlation, mechanism macro-F1 and R2. These controls reduce but do not erase dependence on the controlled answer-pair construction. See Source Data Fig. 7 and the Source Data manifest.}
\end{figure}

\FloatBarrier

\subsubsection*{Is the observed geometry carried by learned vocabulary identity?}
The early readback results were compatible with a strong explanation: learned vocabulary rows might carry the geometry that made the layerwise patterns informative. This interpretation supplied a discriminating prediction. If learned row identity was essential, row permutation or replacement by matched random projections should sharply reduce preservation. We tested the learned output projection against three controls. Row permutation preserved the row set while removing identity. The Gaussian control replaced learned rows within a matched projection-selection pipeline. Random-K disrupted conditional selection.

Learned-W and row-permuted-W readbacks preserved hidden geometry almost identically (Fig. 8b,d). Gaussian projections often preserved more, and the pattern persisted across sampled layers (Fig. 8b,c). Random-K selection sharply reduced preservation (Fig. 8e). These controls reject learned output-row identity as the source of the preserved geometry. They instead identify projection and conditional selection as the operations that retain it. TopK/VIM therefore remains a useful diagnostic readback, but not a privileged mechanism coordinate.

\begin{figure}[!htbp]
\centering
\includegraphics[width=0.98\textwidth,height=0.76\textheight,keepaspectratio]{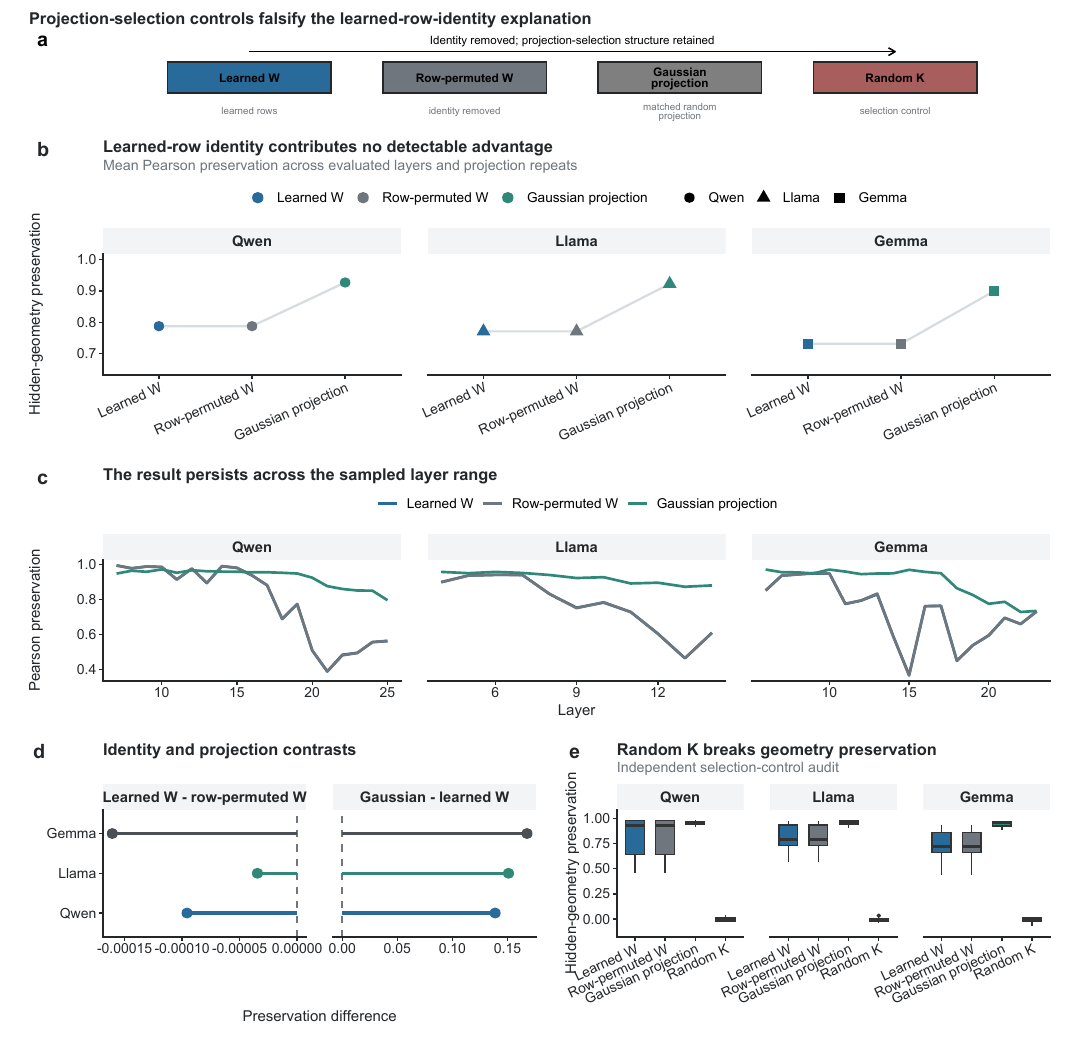}
\caption{Projection-selection controls falsify the learned-row-identity explanation. \textbf{a,} Control logic. The learned output projection was compared with a row-permuted version that removes vocabulary-row identity while preserving the row set, and with a matched Gaussian projection. Random-K selection was evaluated in a separate cross-checkpoint selection-control audit. \textbf{b,} Mean Pearson preservation of pairwise hidden-state geometry across evaluated layers for representative Qwen, Llama and Gemma checkpoints. Learned and row-permuted output projections are indistinguishable at the displayed scale, whereas the Gaussian projection is stronger. \textbf{c,} Layer-resolved preservation averaged over projection repeats. The learned-W and row-permuted-W curves overlap almost exactly; apparent disappearance of the learned-W curve reflects this overlap rather than missing data. \textbf{d,} Model-level contrasts. Learned-W minus row-permuted-W preservation is approximately zero, while Gaussian-minus-learned-W preservation is positive in all three checkpoints. \textbf{e,} Independent controlled proxy-audit dataset selection control. Learned-W, row-permuted-W and Gaussian TopK constructions preserve hidden geometry substantially better than random-K selection. Panel e is not pooled with the projection-selection geometry control projection repeats in panels b-d. Together, the controls reject learned output-row identity as the explanation for readback geometry while retaining vocabulary readbacks as non-random diagnostic observables. Gaussian projection performance does not establish a superior mechanism coordinate. See Source Data Fig. 8 and Source Data manifest.}
\label{fig:8}
\end{figure}

\FloatBarrier

The control removed the basis for privileging vocabulary-facing coordinates. We next compared them directly with variables derived from hidden states.

\subsubsection*{Which object is closer to the process than the vocabulary readback?}
Once learned row identity failed, vocabulary-facing variables could no longer be treated as privileged mechanism coordinates. The failure forced a direct comparative question: did variables derived from hidden states and their ordered changes provide stronger coordinates for later dominance, phase and mechanism labels than the projection-selection readback? The proxy-ranking audit tested this comparison using 960 prompt-condition instances from 80 relation-graph groups and 12 conditions, evaluated by five-fold grouped cross-validation (Fig. 9a). This dataset is distinct from the 96-prompt controlled subset used in the projection-selection audit.

\begin{figure}[!htbp]
\centering
\includegraphics[width=0.98\textwidth,height=0.76\textheight,keepaspectratio]{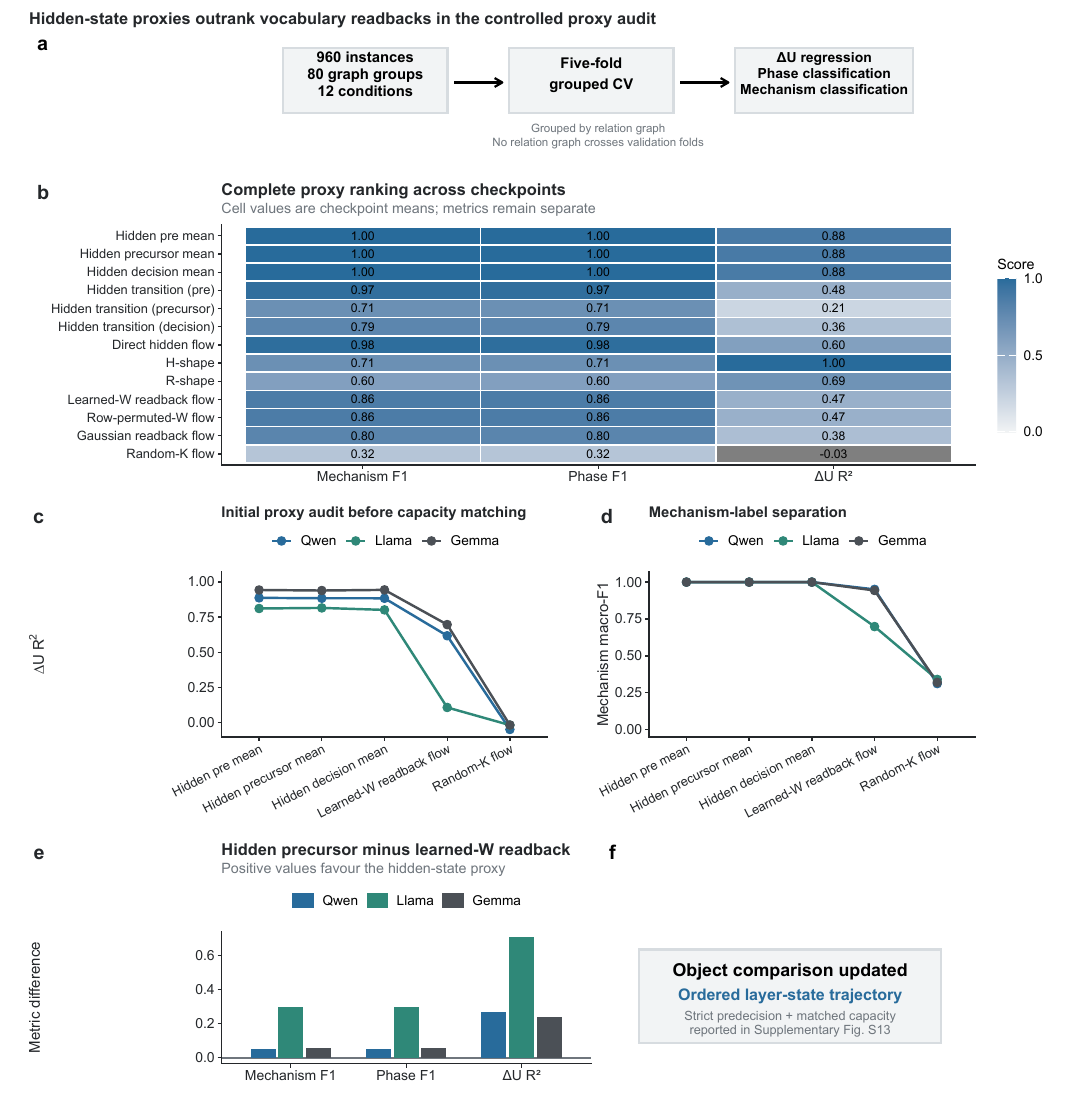}
\caption{Hidden-state proxies outrank vocabulary readbacks in the controlled proxy audit. \textbf{a,} Proxy-audit design for 960 prompt-condition instances from 80 relation-graph groups and 12 conditions, evaluated by five-fold grouped cross-validation. \textbf{b,} Initial proxy ranking across representative Qwen, Llama and Gemma checkpoints; cells report \(\Delta U\) \(R^2\), phase macro-F1 and mechanism macro-F1 separately. \textbf{c,} Checkpoint-specific \(\Delta U\) regression for selected hidden-state, learned-W readback and random-K proxies. \textbf{d,} Mechanism-label macro-F1 for the same proxies. \textbf{e,} Hidden-precursor minus learned-W performance for each metric. \textbf{f,} Object-update decision. Hidden-state variables lead this initial audit; the strict-predecision matched-capacity test in Supplementary Fig. S13 retains their mechanism-label advantage while showing checkpoint- and dimension-dependent \(\Delta U\) results. H-shape is a compressed descriptor, not complete trajectory recovery. See Source Data Fig. 9 and Source Data manifest.}
\label{fig:9}
\end{figure}

\FloatBarrier

Hidden predecision, precursor and decision summaries ranked above learned-W, row-permuted-W and Gaussian readback flows for later \(\Delta U\), phase and mechanism labels in the tested setting (Fig. 9b-d). The hidden precursor advantage over learned-W readback was positive across the three evaluated metrics and all representative checkpoints, with the largest \(\Delta U\) advantage in Llama (Fig. 9e). A second audit asked whether each proxy improved closure of a future-state model. Fold-local dimensionality reduction was fitted inside each grouped training fold. Hidden and trajectory-shape variables produced larger future-state and mechanism-label gains than vocabulary readbacks, whereas random-K was near zero or negative (Fig. 10b-d). A composite gain summary gave the same ordering but was retained only as a secondary view because its component metrics are not directly comparable (Fig. 10e).

\begin{figure}[!htbp]
\centering
\includegraphics[width=0.98\textwidth,height=0.76\textheight,keepaspectratio]{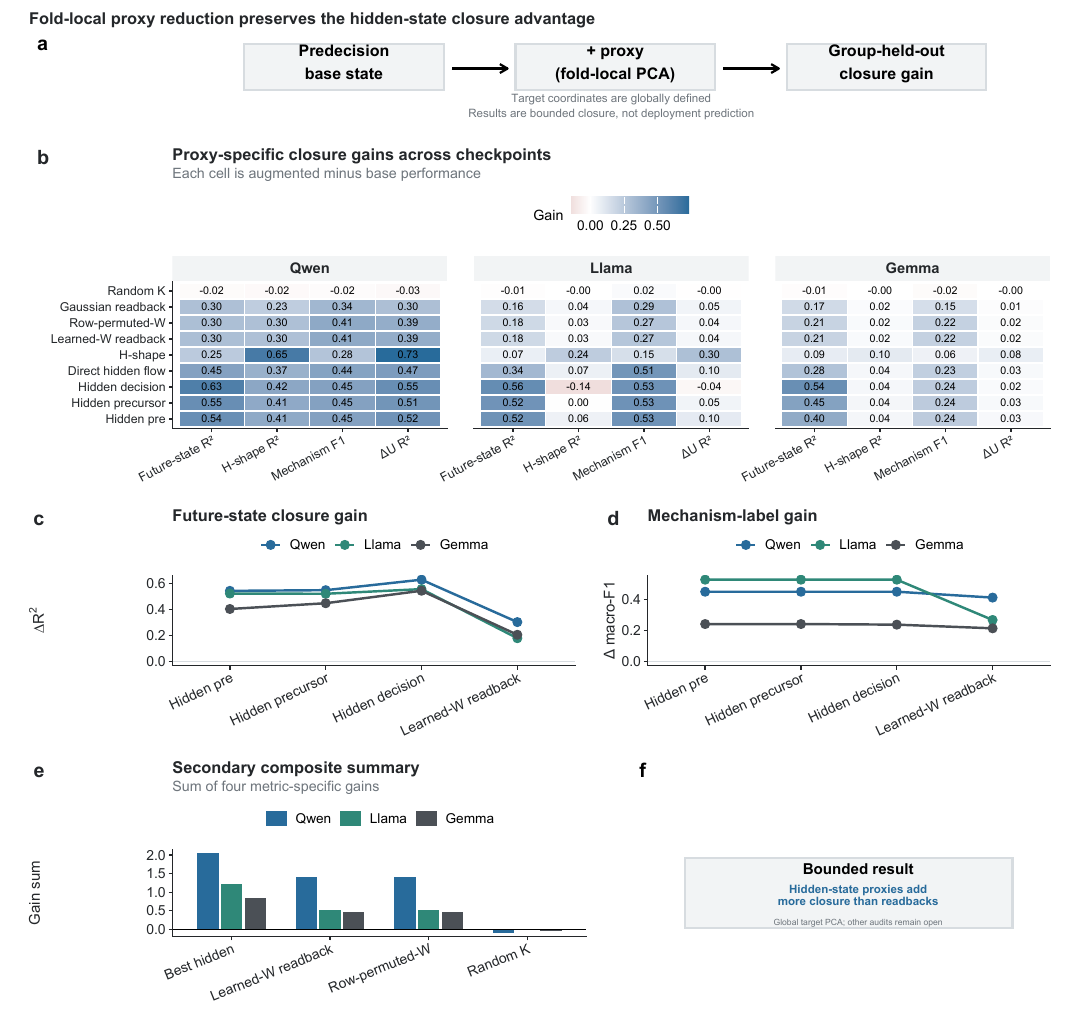}
\caption{Fold-local proxy reduction preserves the hidden-state closure advantage. \textbf{a,} Corrected closure-gain design. Each proxy augments a low-dimensional predecision base state. For proxies exceeding the dimensional cap, PCA is fitted only on the training groups of each five-fold relation-graph-grouped cross-validation split and then applied to held-out groups. \textbf{b,} Augmented-minus-base gains for future-state \(R^2\), H-shape \(R^2\), mechanism macro-F1 and \(\Delta U\) \(R^2\) across representative checkpoints. \textbf{c,d,} Checkpoint-specific gains for selected hidden-state summaries and learned-W readback flow. \textbf{e,} Secondary composite view, defined as the sum of the four displayed metric-specific gains. The composite is not treated as a single physical quantity. Learned-W and row-permuted-W readback gains remain nearly equal (maximum score difference 0.001078), while random-K gains are near zero or negative. \textbf{f,} Boundary. Hidden/state proxies add more group-held-out closure than vocabulary readbacks in this corrected analysis, but the future hidden-state target PCA and \(\Delta U\) orientation are globally defined descriptive coordinates. The result therefore does not establish deployment-style future prediction. Supplementary Fig. S13 separately reports the completed strict-predecision matched-capacity audit. See Source Data Fig. 10 and Source Data manifest.}
\label{fig:10}
\end{figure}

\FloatBarrier

\begin{figure}[!htbp]
\centering
\includegraphics[width=0.98\textwidth,height=0.76\textheight,keepaspectratio]{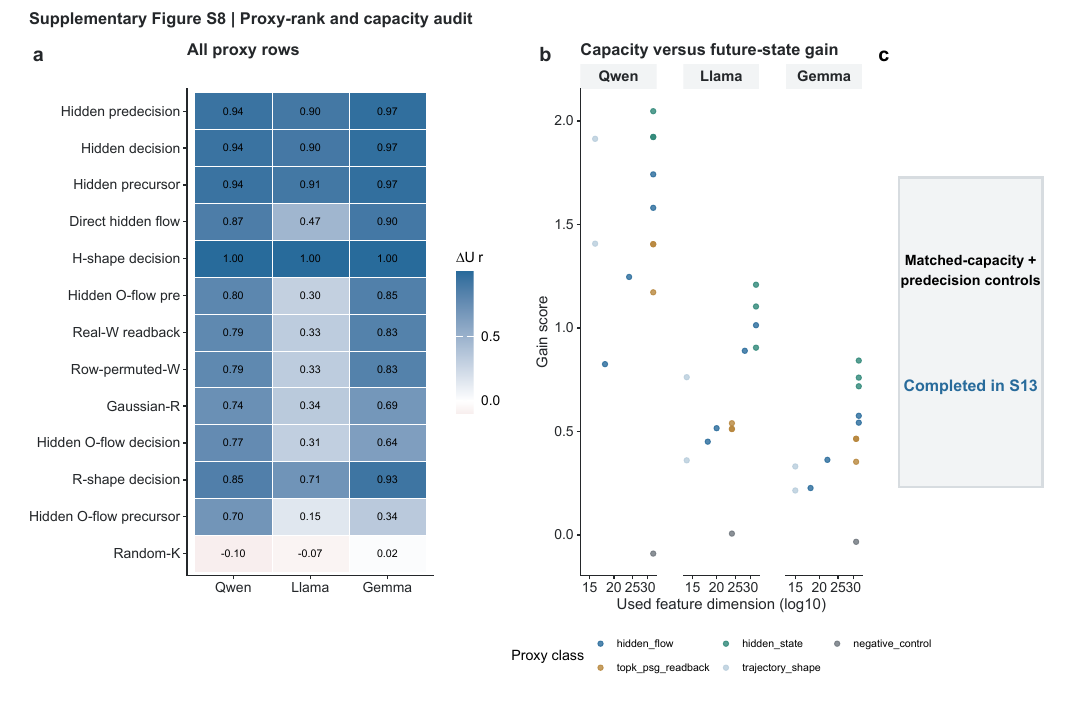}
\caption*{proxy-coordinate proxy-rank and capacity audit. \textbf{a,} Full proxy-by-checkpoint \(\Delta U\) correlation matrix. \textbf{b,} Future-state gain score against retained feature dimension. \textbf{c,} Design of the strict-predecision matched-capacity audit completed in Supplementary Fig. S13. The initial ranking is not interpreted as capacity matched. See Source Data Figs. 9-10 and the Source Data manifest.}
\end{figure}

\FloatBarrier

Together, the grouped comparisons establish the ordered layer-state trajectory as the primary empirical object for the remaining mechanism analyses. We define \(T_{m,p}=(h_{m,p,1},\ldots,h_{m,p,L_m})\) as the ordered hidden states for checkpoint \(m\) and prompt \(p\) (Fig. 11a). From this point onward, mechanism claims refer to this ordered hidden-state trajectory; vocabulary readbacks are retained only as diagnostic projections or comparison coordinates. Checkpoint-specific projections showed organized centroid paths and depth-dependent separation, with much of the variation concentrated in high-variance dimensions (Fig. 11b-e). These projections orient the reader to the trajectory; the quantitative tests use the declared high-dimensional coordinates.

\begin{figure}[!htbp]
\centering
\includegraphics[width=0.98\textwidth,height=0.76\textheight,keepaspectratio]{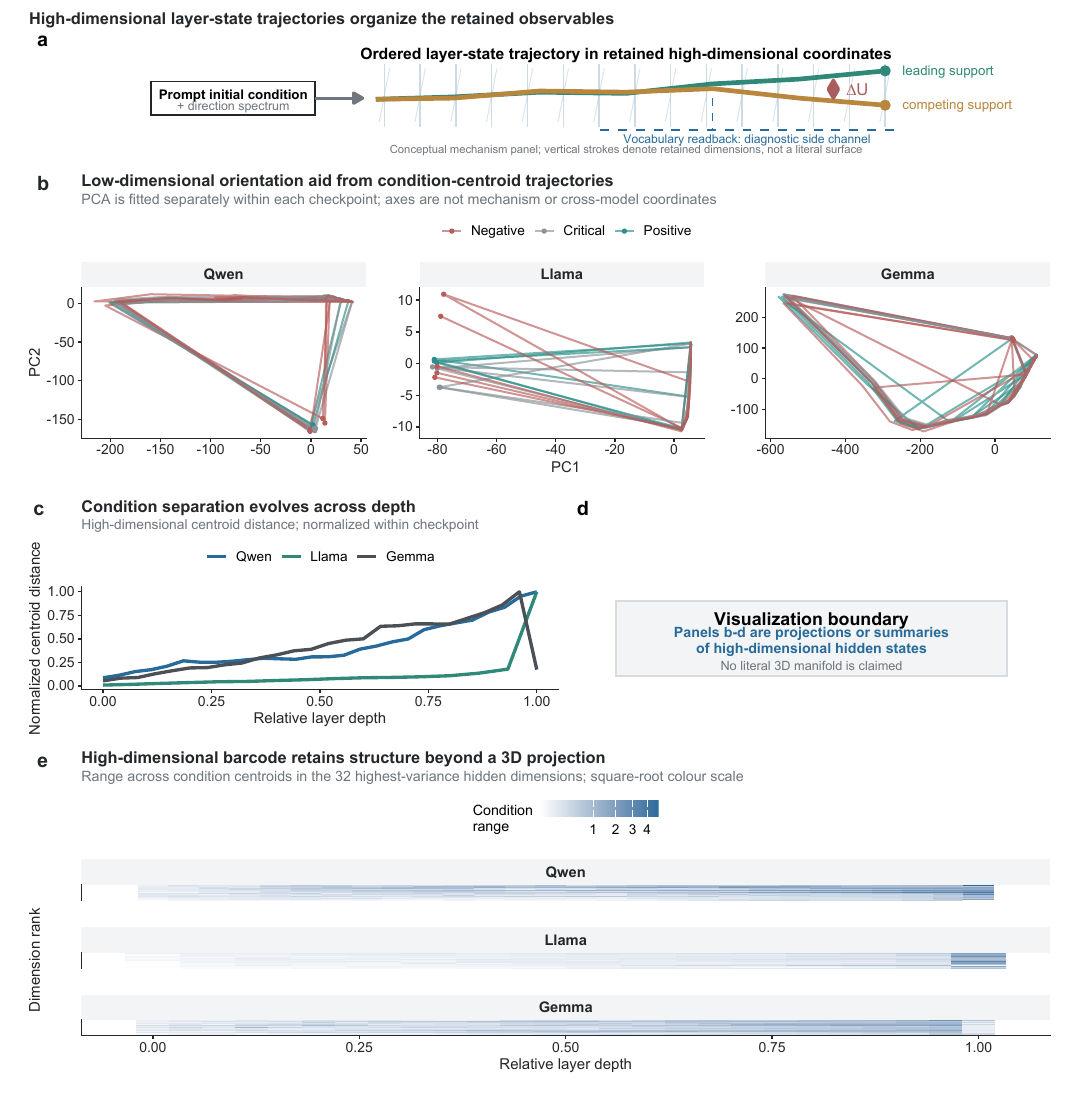}
\caption{High-dimensional layer-state trajectories organize the retained observables. \textbf{a,} Conceptual high-dimensional mechanism view. A prompt supplies the initial condition and direction spectrum; the ordered layer-state sequence develops in retained high-dimensional coordinates, while leading and competing support separate along the bounded observable \(\Delta U\). Vertical strokes denote retained dimensions rather than a literal surface. Vocabulary-facing readback is shown as a diagnostic side channel. The ordered hidden-state sequence remains the primary empirical object; transition coordinates, \(\Delta U\) and structured residuals are derived diagnostics. \textbf{b,} Low-dimensional orientation aid obtained from the full hidden-state arrays for 960 prompt-condition instances. PCA was fitted separately within each checkpoint over all condition-by-layer centroids. Lines connect successive layers and are coloured by controlled phase label. Axes are checkpoint-specific projection coordinates, not mechanism coordinates, and cannot be compared numerically across panels. \textbf{c,} Mean pairwise Euclidean distance among high-dimensional condition centroids at each layer, normalized by the within-checkpoint maximum. Late-layer decreases for Qwen and Gemma are present in the extracted data and are not missing segments. \textbf{d,} Visualization boundary: panels b-e are projections or summaries of high-dimensional hidden states and do not claim that the underlying object is literally a three-dimensional manifold. \textbf{e,} Range across condition centroids in the 32 hidden dimensions with highest variance in each checkpoint. Colour uses a square-root scale to reveal non-negative range structure without altering the underlying values. Dimension ranks and PCA axes are selected separately within each checkpoint. See Source Data Fig. 11 and Source Data manifest.}
\label{fig:11}
\end{figure}

\FloatBarrier

Dimensional capacity and proximity to the target could still explain that ranking. We therefore repeated the comparison using only predecision layers and reduced every proxy to 2, 4, 6 or 8 dimensions inside each grouped training fold. Hidden coordinates retained higher mechanism-label macro-F1 at every checkpoint and dimension (Supplementary Fig. S13b,c). Their \(\Delta U\) advantage was not universal: it was positive throughout Qwen, near zero at intermediate dimensions in Llama, and reversed at two dimensions in Gemma before becoming positive at 4-8 dimensions (Supplementary Fig. S13a,c). The matched audit supports hidden trajectories for controlled mechanism-label organization while showing that dominance prediction depends on checkpoint and retained capacity. We next used transition closure to examine how successive states were related.

\begin{figure}[!htbp]
\centering
\includegraphics[width=0.98\textwidth,height=0.76\textheight,keepaspectratio]{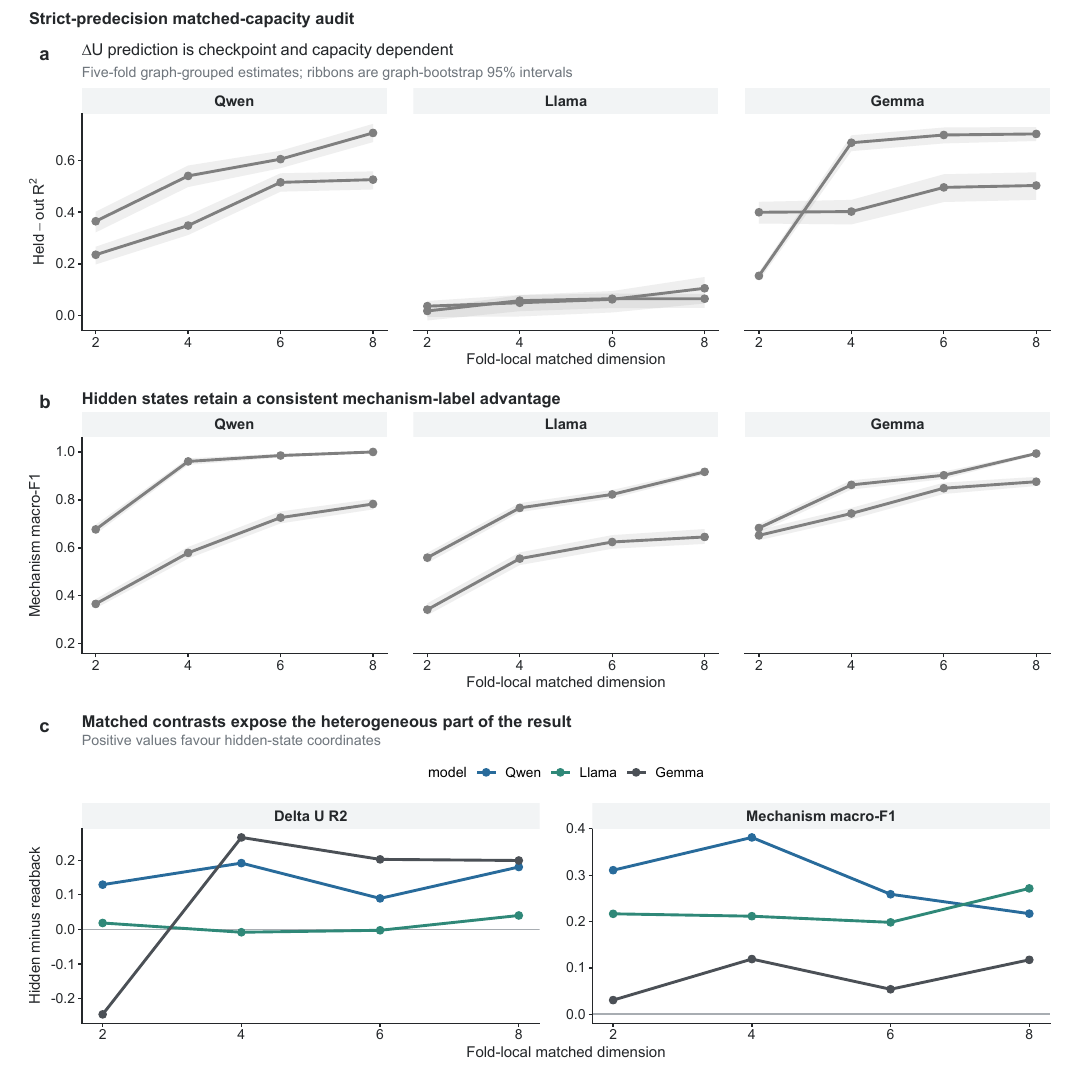}
\caption*{Strict-predecision matched-capacity object audit. \textbf{a,} Held-out \(\Delta U\) \(R^2\) for hidden-state and vocabulary-readback coordinates after both were reduced to 2, 4, 6 or 8 dimensions inside each graph-grouped training fold. Only transitions before the declared decision window entered either proxy. Ribbons show graph-bootstrap 95\% intervals. The hidden-state advantage is checkpoint and capacity dependent: it is positive throughout Qwen, near zero at intermediate dimensions in Llama, and reverses at two dimensions in Gemma. \textbf{b,} Mechanism-label macro-F1 under the same strict-predecision, matched-capacity construction. Hidden-state coordinates are higher at every tested checkpoint and dimension. \textbf{c,} Hidden-minus-readback contrasts, shown without aggregation across metric types. Positive values favour hidden-state coordinates. The result supports a stable mechanism-label advantage but not a universal \(\Delta U\) advantage. Five-fold cross-validation kept relation graphs intact; uncertainty was estimated by 500 graph-level bootstrap resamples of out-of-fold predictions. See Source Data Supplementary Fig. S13 and the Source Data manifest.}
\end{figure}

\FloatBarrier

\section{Following the ordered hidden-state trajectory}
\subsubsection*{Can layer-to-layer evolution be described without claiming prediction?}
A trajectory object by itself leaves the relation between successive states unspecified. We therefore asked whether compact coordinates of the observed transition could close more of the adjacent-state relation than state descriptors alone. The crucial distinction was whether such closure remained descriptive or could be recovered from information available before the transition. The six-feature H-shape descriptor, which records local velocity changes, reversals and centre backtracking, modestly increased closure over state alone, whereas continuous latent transition coordinates \(O_l^{cont}\) produced a much larger increase. Macro \(R^2\) rose from 0.820 for the state model and 0.824 for state plus H-shape to 0.980 for state plus transition coordinates and 0.993 for a bilinear state-transition model (Fig. 12b). Removing \(O_l^{cont}\) caused the largest ablation drop (Fig. 12c).

\begin{figure}[!htbp]
\centering
\includegraphics[width=0.98\textwidth,height=0.76\textheight,keepaspectratio]{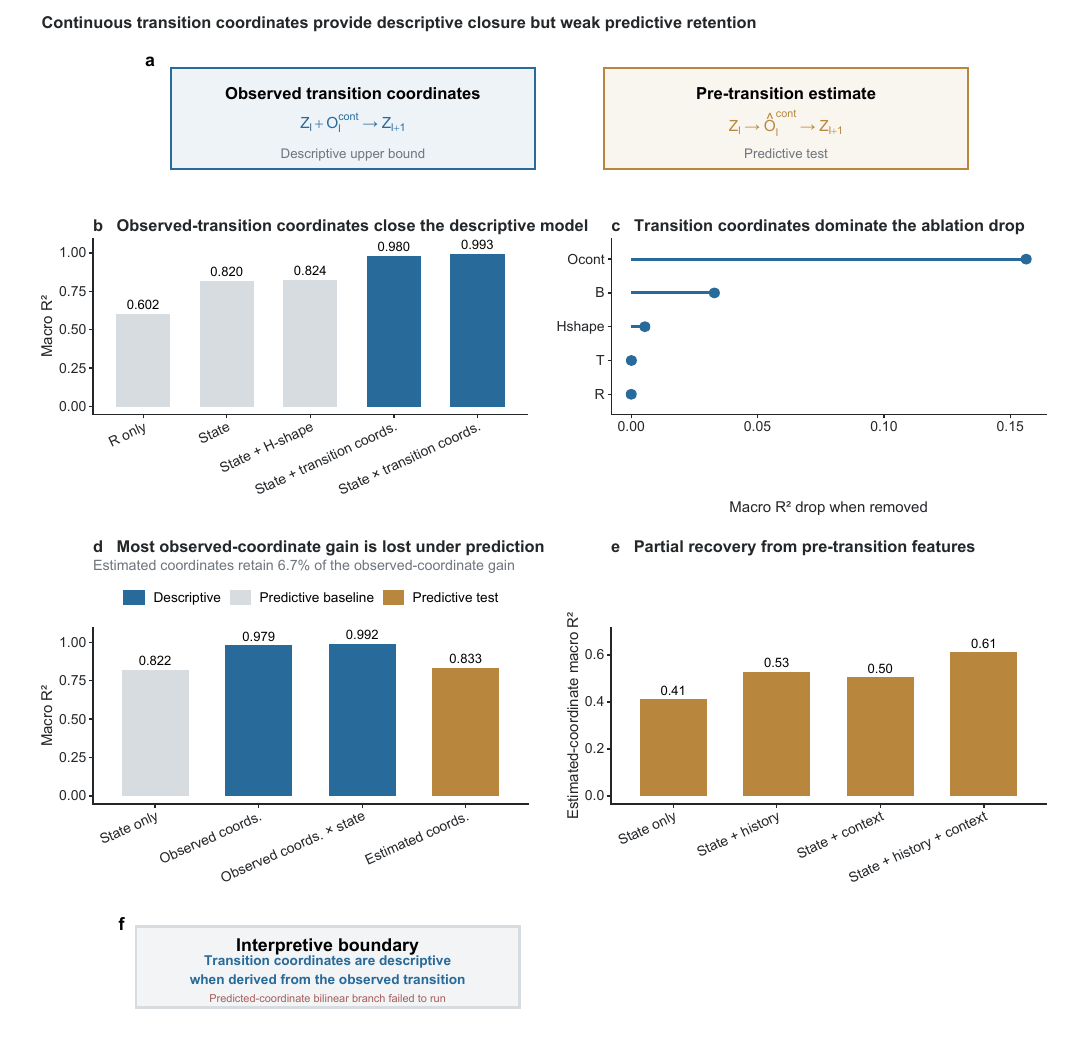}
\caption{Continuous latent transition coordinates provide descriptive closure but weak predictive retention. \textbf{a,} Information boundary. Observed continuous transition coordinates \(O_l^{cont}\) are derived from the realized transition and provide a descriptive upper bound. Estimated coordinates \(\hat O_l^{cont}\) use pre-transition state/history/context and constitute the predictive test. \textbf{b,} one-step transition-closure audit model ladder. Adding observed \(O_l^{cont}\) raises grouped macro-\(R^2\) from 0.824 for state plus H-shape to 0.980; state-coordinate interactions reach 0.993. \textbf{c,} Leave-component-out analysis. Removing \(O_l^{cont}\) produces the largest macro-\(R^2\) drop. \textbf{d,} transition-coordinate-7F comparison. State-only prediction reaches 0.822, observed coordinates reach 0.979, and the best model using pre-transition estimated coordinates reaches 0.833. Estimated coordinates therefore retain 6.7\% of the observed-coordinate gain under the implemented test. \textbf{e,} Predictability of \(O_l^{cont}\) from increasingly rich pre-transition feature sets; the best macro-\(R^2\) is 0.61. \textbf{f,} Interpretation boundary. Adjacent-state-derived \(O_l^{cont}\) supports descriptive reconstruction but not an equivalent predictive claim. The predicted-coordinate bilinear branches in transition-coordinate-7F terminated with implementation errors and are excluded rather than imputed. See Source Data Fig. 12 and Source Data manifest.}
\label{fig:12}
\end{figure}

\FloatBarrier

Observed transition coordinates improved closure, but they used information from the transition being reconstructed. A pre-transition estimator recovered only part of that information. The best estimated-coordinate model reached macro \(R^2=0.833\), retaining 6.7\% of the observed-coordinate gain over the state-only baseline (Fig. 12d). History and context helped estimate \(O_l^{cont}\), although macro \(R^2\) remained between 0.41 and 0.61 in the tested variants (Fig. 12e). We therefore use \(O_l^{cont}\) as a descriptive coordinate unless it is estimated before the target transition. This left open whether the ordered states followed a simpler global route.

\begin{figure}[!htbp]
\centering
\includegraphics[width=0.98\textwidth,height=0.76\textheight,keepaspectratio]{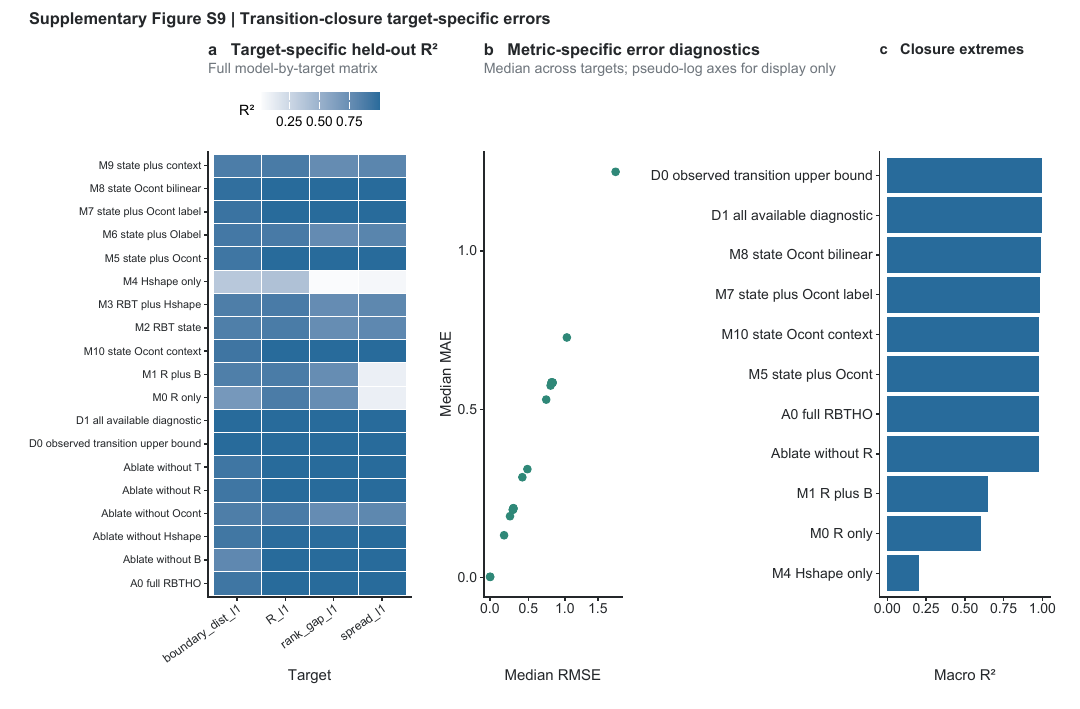}
\caption*{transition-coordinate target-specific errors. \textbf{a,} Held-out model-by-target R2 matrix. \textbf{b,} Median RMSE and MAE across targets; values are metric specific and not directly comparable across metric types. \textbf{c,} Highest and lowest model-ladder macro-R2 values; the complete ladder is retained in Source Data. Observed-transition O\_cont is descriptive unless estimated without target-layer information. See Source Data Fig. 12 and the Source Data manifest.}
\end{figure}

\FloatBarrier

\subsubsection*{Are ordered layer-state trajectories close to an ideal shortest route?}
Compact transition coordinates left open whether the observed route through layer-state space had a simple geometric form. We proposed the strict conjecture that ordered trajectories approximated an ideal geodesic relative to the chord joining the first and last analysed states. The order-shuffle control retained the same states and endpoints while permuting only the interior layer order; alignment, curvature and detour were then recomputed. The conjecture had two levels of prediction: real layer order should be more aligned and less curved or detoured than these shuffles, and an exact account should leave little residual motion. Real trajectories satisfied the first level: they were more aligned with the reference direction than shuffles, although the gain was heterogeneous across checkpoints (Fig. 13b), and curvature and detour were lower across the principal window and K comparisons (Fig. 13c,d and Supplementary Fig. S10).

\begin{figure}[!htbp]
\centering
\includegraphics[width=0.98\textwidth,height=0.76\textheight,keepaspectratio]{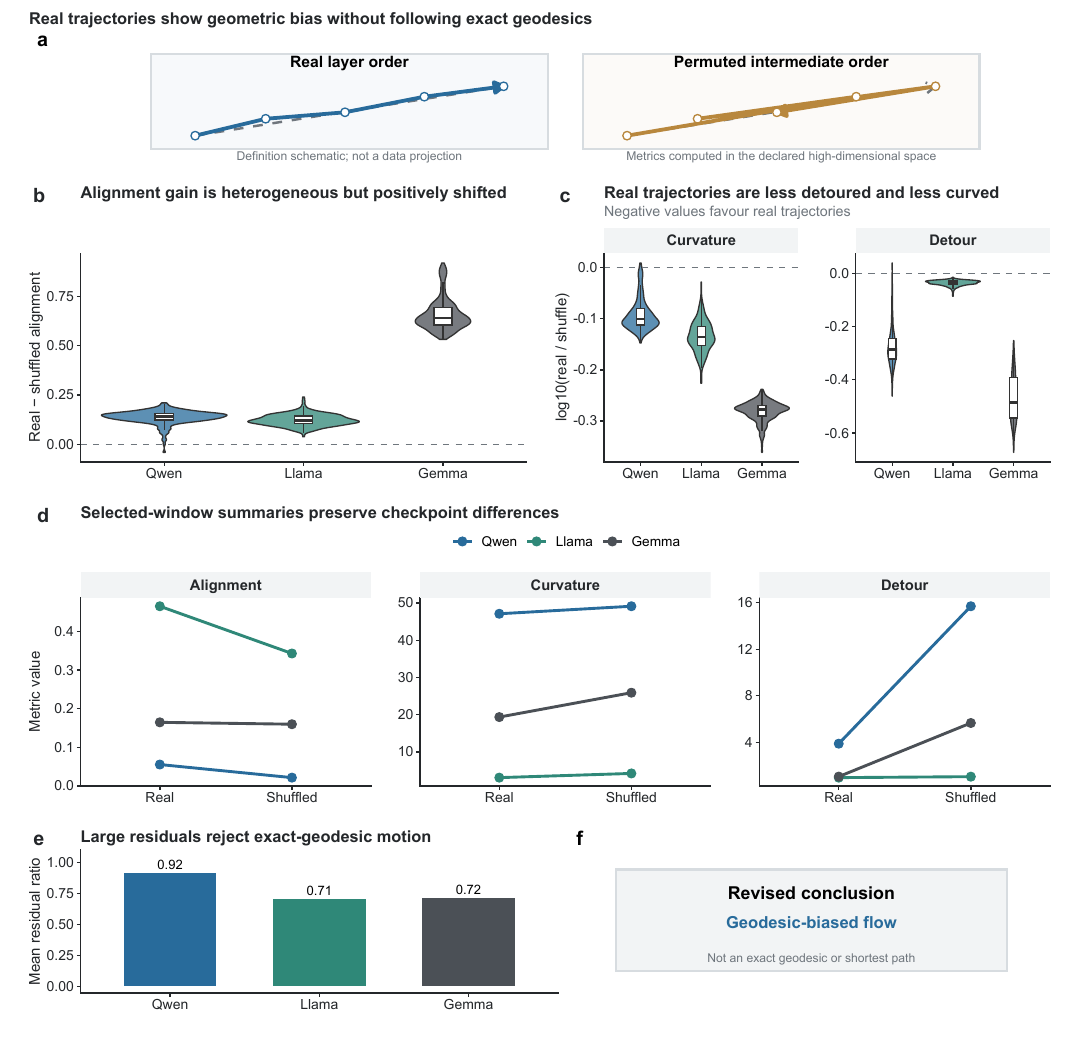}
\caption{Real trajectories show geometric bias without following exact geodesics. \textbf{a,} Control schematic comparing real layer order with an interior-layer permutation relative to a declared endpoint direction. The panel explains the test; real-data trajectory projections appear in Fig. 11b. \textbf{b,} Decision-window alignment gain for 1,296 checkpoint-prompt observations at \(K=500\), defined as real minus shuffled mean alignment. \textbf{c,} Detour and curvature contrasts expressed as \(\log_{10}(\mathrm{real}/\mathrm{shuffle})\); negative values favour real layer order. \textbf{d,} Real and shuffled values in the separately selected best-bias window for each checkpoint. \textbf{e,} Mean real-trajectory residual ratio. This norm ratio is not a percentage of explained variance. Its large value excludes exact-geodesic motion, whereas geometric bias is established by the paired contrasts in panels b-d. \textbf{f,} The retained evidence supports geodesic-biased flow relative to the declared order shuffles, not exact geodesics. See Source Data Fig. 13 and Source Data manifest.}
\label{fig:13}
\end{figure}

\FloatBarrier

\begin{figure}[!htbp]
\centering
\includegraphics[width=0.98\textwidth,height=0.76\textheight,keepaspectratio]{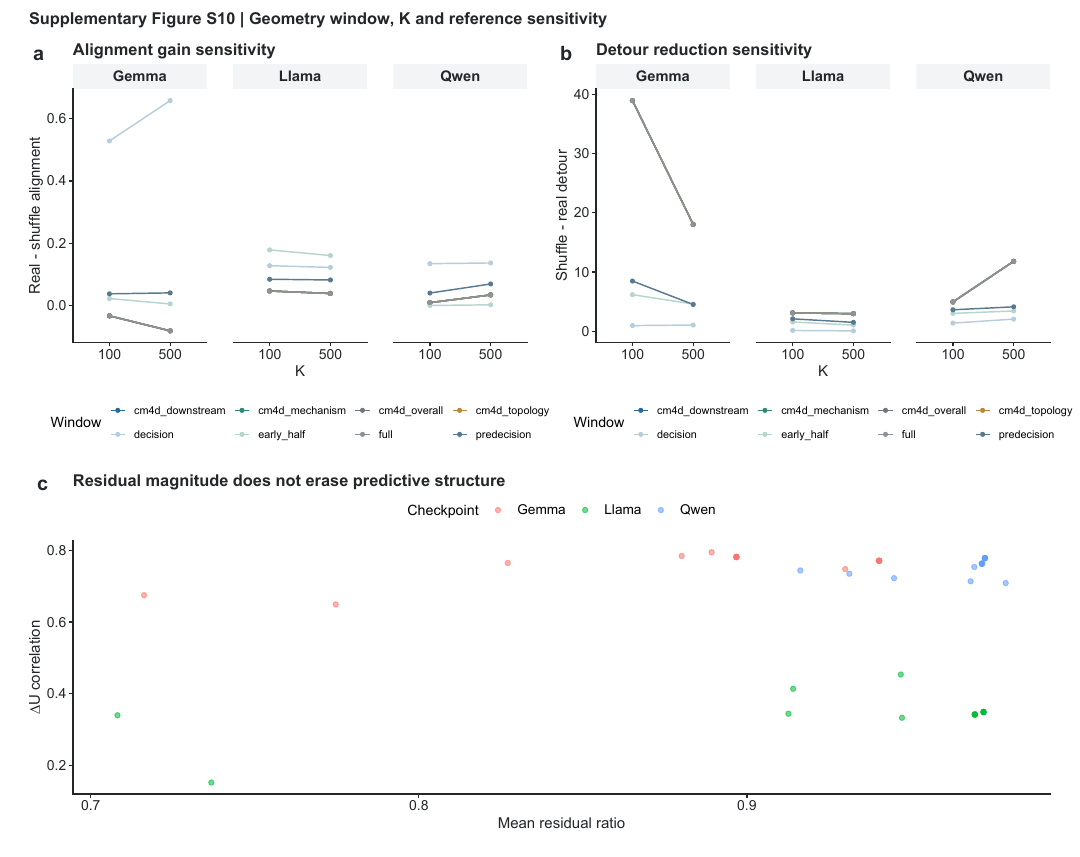}
\caption*{Geometry window, K and reference sensitivity. \textbf{a,} Real-minus-shuffle alignment by window and K. \textbf{b,} Shuffle-minus-real detour by window and K. \textbf{c,} Residual ratio against \(\Delta U\) correlation. The retained claim is geodesic-biased flow, not exact geodesic following. See Source Data Fig. 13 and the Source Data manifest.}
\end{figure}

\FloatBarrier

Exact-geodesic dynamics were decisively excluded. Mean residual ratios were 0.92 for Qwen, 0.71 for Llama and 0.72 for Gemma (Fig. 13e), leaving substantial motion outside the reference direction. These norm ratios are not percentages of explained variance and do not measure the significance of the aligned component. Geometric bias is established by the paired real-versus-order-shuffle contrasts in alignment, detour and curvature. Real trajectories were consistently more organized under those comparisons despite their large residuals. We call this combination geodesic-biased flow: directional organization without exact geodesic following. The residual then became a second empirical object rather than an error term to discard.

\subsubsection*{Are deviations from the reference direction signal or noise?}
Once exact-geodesic motion was rejected, the unexplained component admitted two competing interpretations: nuisance deviation around an organized direction, or a second source of structured information. Each observed transition was summarized by a reference-aligned proxy and an off-direction residual proxy; the two feature families were evaluated separately and together under the same grouped splits. This is an implemented proxy decomposition, not an exhaustive orthogonal decomposition of hidden-state space. Under the noise-only account, residual features should add no stable information beyond the reference-aligned family. Residual-only features instead retained \(\Delta U\) information across windows and checkpoints (Fig. 14b). Combined projection and residual features generally matched or exceeded either family alone, although effect size and the best feature family varied by checkpoint (Fig. 14b,c). The two proxy families were likewise complementary for \(\Delta U\) and mechanism-label separation (Fig. 14d,e).

\begin{figure}[!htbp]
\centering
\includegraphics[width=0.98\textwidth,height=0.76\textheight,keepaspectratio]{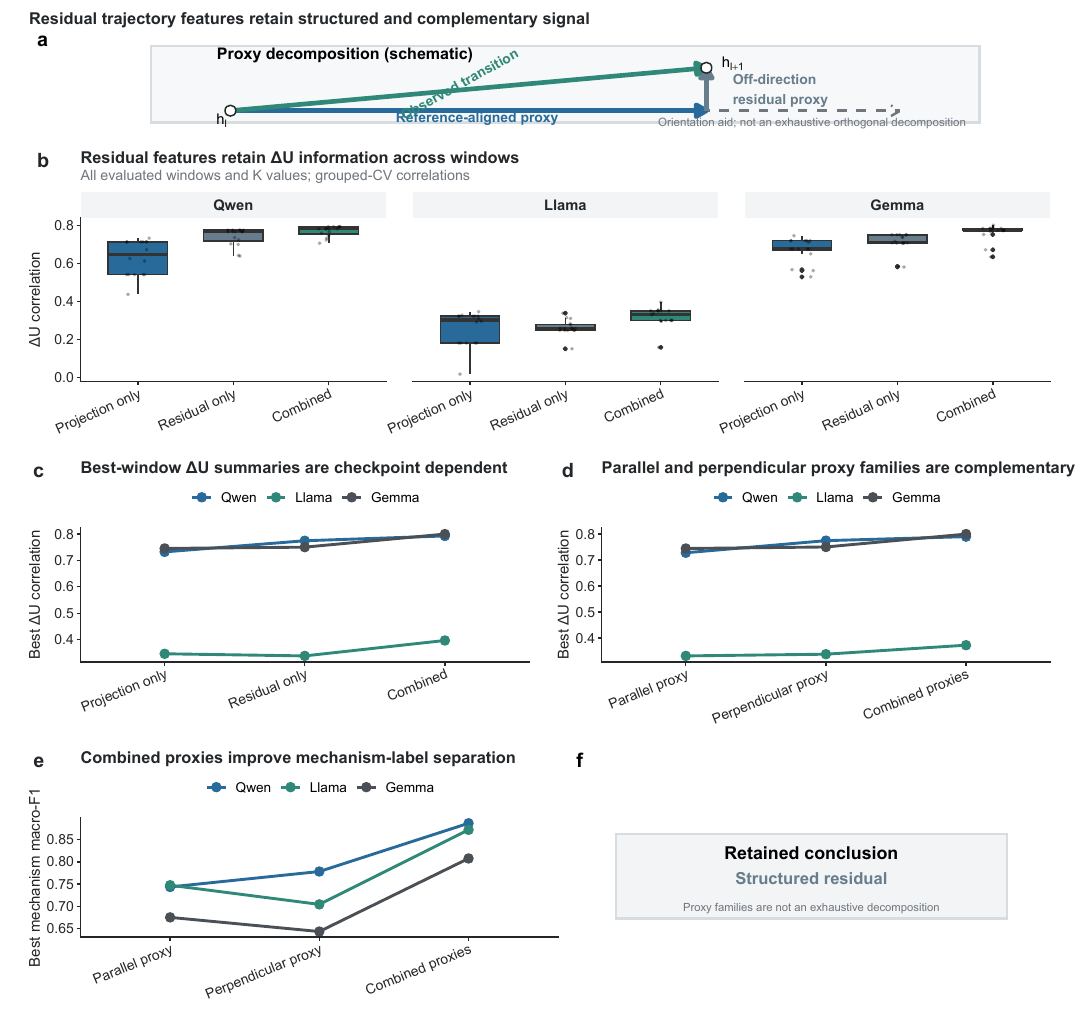}
\caption{Residual trajectory features retain structured and complementary signal. \textbf{a,} Definition schematic for the residual-feature analyses feature logic. An observed transition is represented by a reference-aligned proxy component and an off-direction residual proxy; the vector drawing is an orientation aid. Reference-aligned, residual and combined feature families are evaluated separately. These are implemented proxy families rather than an exhaustive orthogonal decomposition of the full hidden-state space. \textbf{b,} Grouped-cross-validation correlation with \(\Delta U\) for projection-only, residual-only and combined feature sets over all declared window labels and \(K\) values. Residual-only signal is present across checkpoints, but Llama provides a counterexample to any claim that residual features always outperform projection features. \textbf{c,} Best-window \(\Delta U\) correlations from projection-residual feature audit. Combined features are strongest in each checkpoint summary, with substantial checkpoint heterogeneity. \textbf{d,} Best \(\Delta U\) correlations for parallel, perpendicular and combined proxy families in parallel-perpendicular proxy audit. \textbf{e,} Mechanism-label macro-F1 for the same proxy families. Combining the two families improves the best summary in all three checkpoints. \textbf{f,} Retained conclusion. Residual features are structured rather than disposable noise, while their stable taxonomy remains open. See Source Data Fig. 14 and Source Data manifest.}
\label{fig:14}
\end{figure}

\FloatBarrier

These results reject the noise-only account of off-direction motion. Residual features carried stable information beyond the reference-aligned family, and the combined representation was generally strongest. Exploratory correction-and-rotation tests found more consistent correction-related measures than rotation measures (Supplementary Fig. S11). The present decomposition does not exhaust the residual, but it establishes that the non-geodesic component is structured. The final experiment asked whether the retained coordinates were also operationally discriminating.

\begin{figure}[!htbp]
\centering
\includegraphics[width=0.98\textwidth,height=0.76\textheight,keepaspectratio]{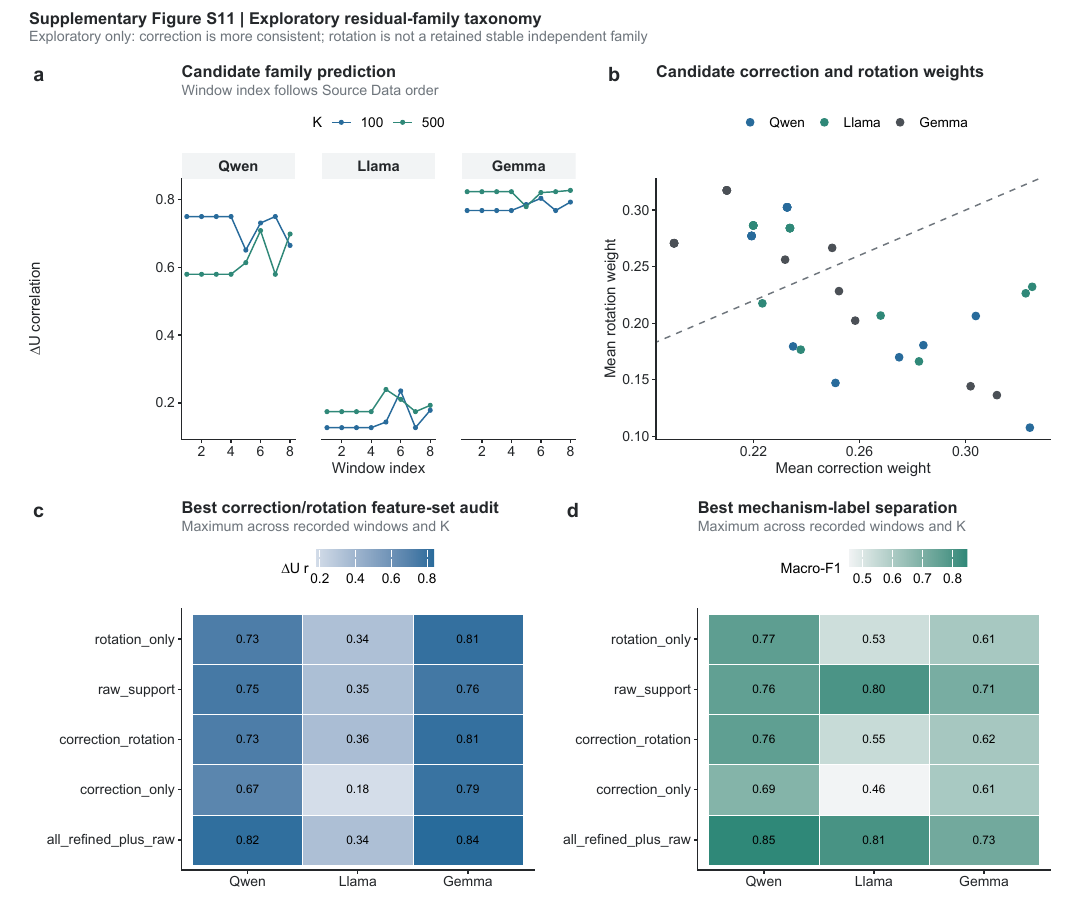}
\caption*{Exploratory residual-family taxonomy. \textbf{a,} Candidate family prediction across windows and K. \textbf{b,} Candidate correction and rotation weights. \textbf{c,d,} Best \(\Delta U\) correlation and mechanism macro-F1 across recorded windows and K for correction/rotation feature sets. This taxonomy is exploratory: correction is more consistent, whereas rotation is not retained as a stable independent family. See Source Data Supplementary Fig. S11 and the Source Data manifest.}
\end{figure}

\FloatBarrier

\section{Testing a bounded operational consequence}
\subsubsection*{Can the revised coordinates support specific internal modulation?}
The preceding analyses were passive measurements. We next asked a narrower operational question: could the retained coordinates select actions that changed the internal endpoint more specifically than fixed and matched-shuffle alternatives? Mechanism and policy models were fitted on training relation graphs, used to select actions for held-out graphs, and compared with a fixed action and three label-shuffle null families. No held-out graph contributed to coordinate fitting, action-label construction or classifier training. The endpoint, specificity, was defined as the target absolute \(\Delta U\) shift minus the non-target absolute shift. Classifier accuracy and macro-F1 values are training diagnostics, whereas specificity was evaluated on held-out graphs.

Mechanism-aware policies exceeded fixed action combinations in all three representative checkpoints (Fig. 15b), but the raw specificity gain was much smaller in Llama than in Gemma. The mechanism labels were assigned from the controlled prompt conditions before fitting; they were not discovered from generated answers or inferred model behaviour. Against the joint mechanism-and-policy shuffle, observed specificity yielded \(z=4.38\) for Qwen, \(4.23\) for Llama and \(4.02\) for Gemma, and exceeded the empirical 95th percentile in each checkpoint (Fig. 15c,d). Results varied across null families. Llama fell below the 95th percentile of the mechanism-policy-pairing shuffle considered alone. All checkpoints exceeded the policy-label and joint-shuffle controls (Supplementary Fig. S12a,b). Policy actions were also imbalanced, with the 0/0 action common in every checkpoint (Supplementary Fig. S12c).

\begin{figure}[!htbp]
\centering
\includegraphics[width=0.98\textwidth,height=0.76\textheight,keepaspectratio]{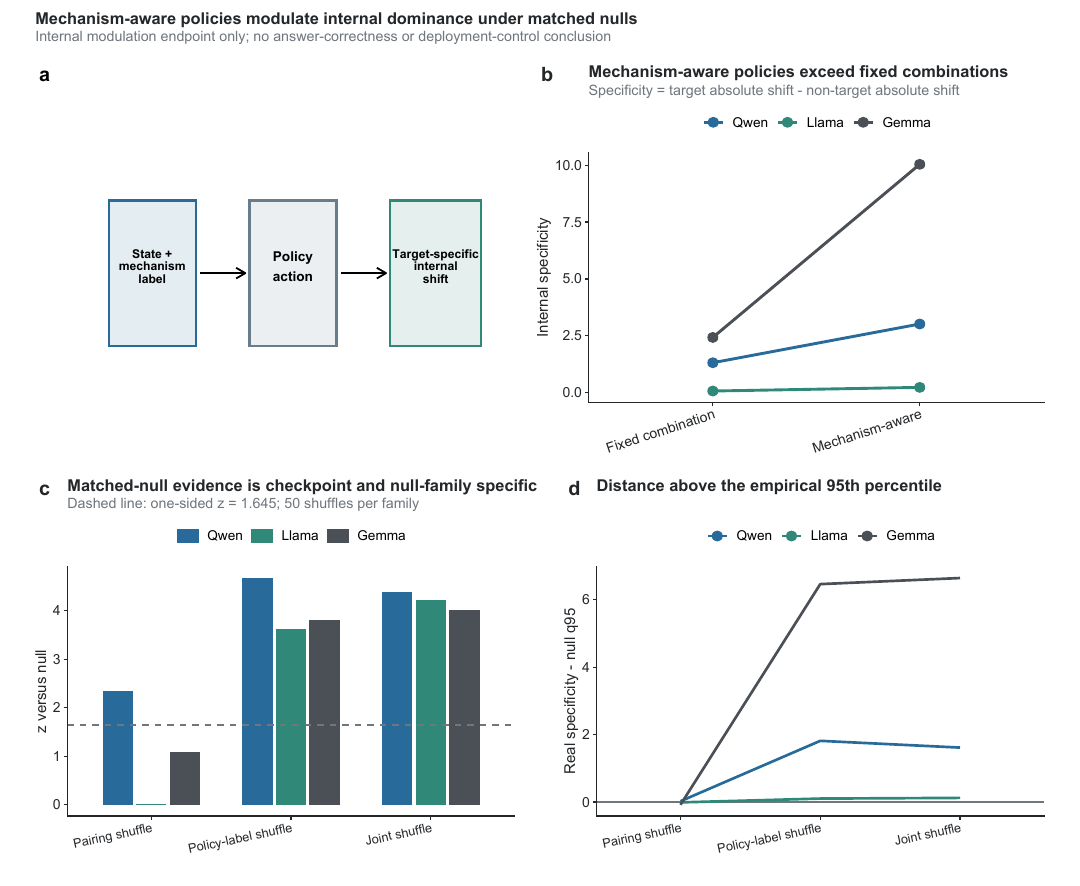}
\caption{Mechanism-aware policies modulate internal dominance under matched nulls. \textbf{a,} Audit workflow: checkpoint state and mechanism label determine an internal policy action, evaluated by a target-specific internal-shift endpoint. \textbf{b,} Internal specificity (closure-condition absolute shift minus non-closure absolute shift) for fixed-combination and mechanism-aware policies. Lines join results within each representative checkpoint. In the action labels, none denotes 0/0 operator/order-precursor weights. \textbf{c,} Standardized separation of the observed mechanism-aware specificity from three 50-shuffle null families. The dashed line marks a one-sided normal-reference threshold of 1.645 and is shown only as a visual guide; empirical quantiles are reported in panel d. \textbf{d,} Difference between observed specificity and the empirical 95th percentile of each matched null. Qwen, Llama and Gemma denote representative checkpoints from the respective families. Mechanism and policy classifier values used in this experiment are training diagnostics, not held-out generalization estimates. The endpoint establishes internal modulation specificity only; it does not establish answer correctness, behavioural benefit or deployment control. See Source Data Fig. 15 and the Source Data manifest.}
\label{fig:15}
\end{figure}

\FloatBarrier

\begin{figure}[!htbp]
\centering
\includegraphics[width=0.98\textwidth,height=0.76\textheight,keepaspectratio]{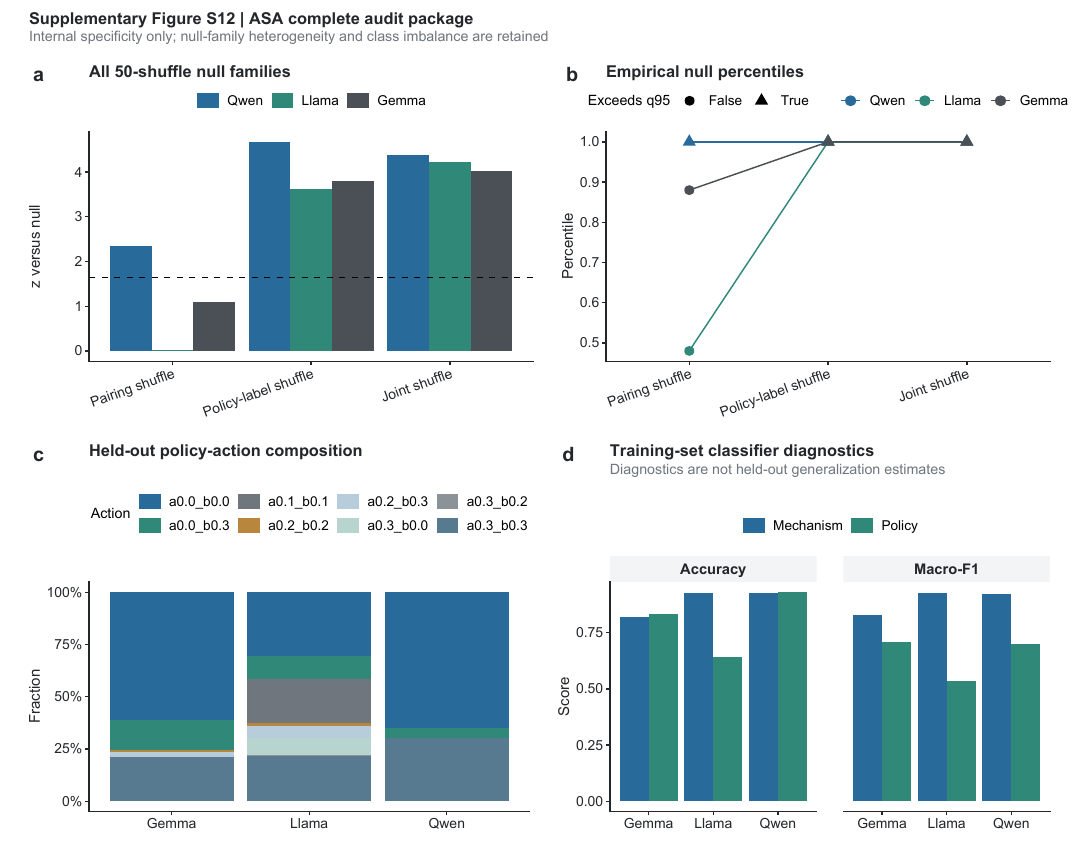}
\caption*{internal-modulation complete audit package. \textbf{a,b,} Standardized and empirical-percentile results for all three 50-shuffle null families. \textbf{c,} Held-out policy-action composition; \protect\path{a0.0_b0.0} denotes 0/0 operator/order-precursor weights. \textbf{d,} Training-set mechanism and policy classifier diagnostics, which are not held-out generalization estimates. Null-family heterogeneity and action imbalance are retained. The endpoint concerns internal modulation only, not answer correctness or deployment control. See Source Data Fig. 15, Source Data Supplementary Fig. S12 and the Source Data manifest.}
\end{figure}

\FloatBarrier

These experiments close a controlled observation-to-operation loop. Trajectory-informed policies produced target-specific internal changes that fixed and label-shuffled alternatives did not reproduce. The result establishes operational relevance for the retained coordinate within the internal dominance endpoint. Whether this modulation changes answer selection or correctness is a separate behavioural question that was not tested here.

\section{Independent controls and output-boundary extension}
\subsubsection*{Does the order effect transfer beyond the relation-graph task?}
The order effect transferred to a separate lexical-category task containing 64 problems and six matched conditions. Real-order alignment exceeded every one of 50 endpoint-preserving shuffles in each checkpoint, with null-standardized effects of 5.27, 6.56 and 6.26 for Qwen, Llama and Gemma (Fig. 16a). A new coordinate fitted only on 44 training problems gave held-out mechanism macro-F1 values of 0.537, 0.562 and 0.591; it exceeded the shuffled-label 95th percentile in Qwen and Llama but not Gemma (Fig. 16b). Clean-condition candidate selection was 0.891, 0.547 and 0.953, making the Llama coordinate result a weak-behaviour boundary. The finite-path geometry therefore transferred more strongly than the one-dimensional coordinate construction. The numerical relation-graph $\Delta U$ axis was not transferred.

\subsubsection*{Is chord-directed order learned during task training?}
Three full random initializations of the same Qwen architecture, evaluated on the same 96 relation-graph prompts, had alignment gains of 0.0794, 0.0803 and 0.0781, all larger than the trained checkpoint's 0.0334 (Fig. 16c). Chord-directed order is therefore not evidence of training-specific inference organization by itself. It is compatible with an ordered architectural scaffold on which task-conditioned separation and intervention must supply the functional evidence. Because the control covered only the Qwen architecture, it does not establish a transformer-wide architectural principle.

\subsubsection*{Does internal trajectory displacement reach the candidate boundary?}
Under the original action budget, closure candidate crossings occurred in 0 of 87 Qwen prompts, 9 of 45 Llama prompts and 6 of 56 Gemma prompts. Cross-fitted maps from $\Delta U$ displacement to candidate-margin displacement reached pooled $R^2=0.850$, and predicted post-intervention margin separated the 15 crossings with AUROC 0.976 (Fig. 16d). Qwen moved 86 of 87 trajectories towards the clean candidate and shifted its mean clean-minus-conflict margin by 2.49 logits, but its initial deficit was larger and 94\% of closure-layer actions saturated at the maximum original action. The zero-crossing result therefore identified an exhausted actuator budget, not absence of internal trajectory control.

\begin{figure}[!htbp]
\centering
\includegraphics[width=0.98\textwidth,height=0.78\textheight,keepaspectratio]{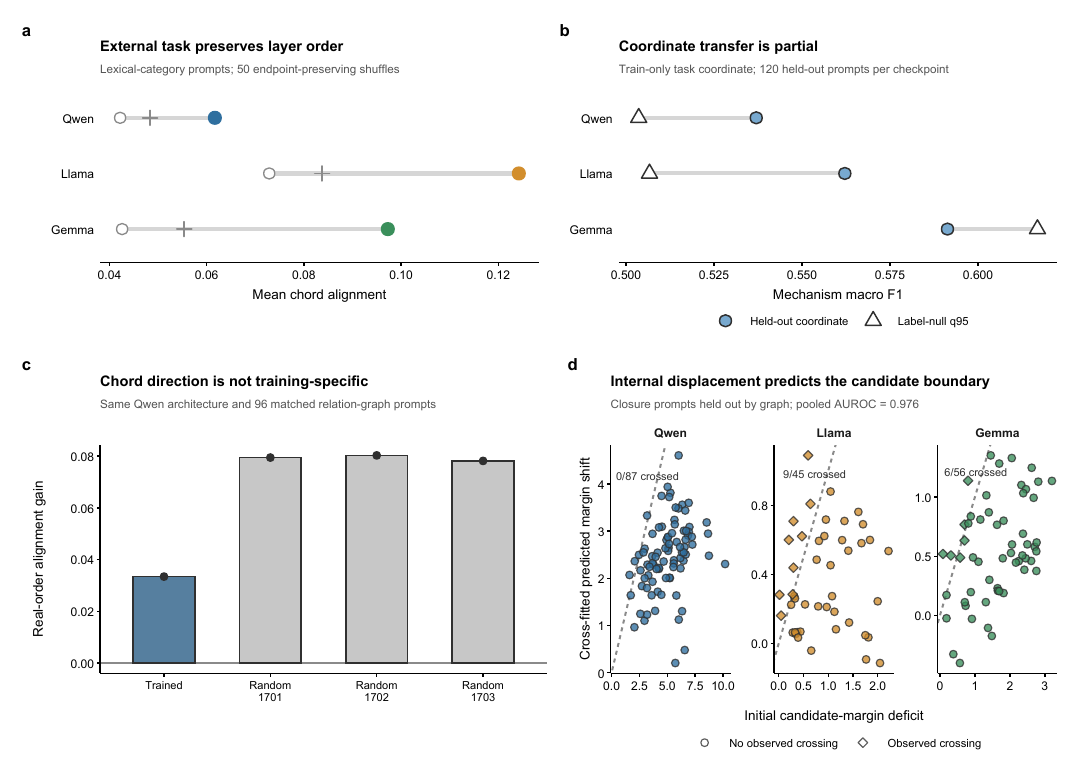}
\caption{Independent task, initialization and boundary controls. \textbf{a,} Endpoint-preserving layer-order nulls in a separate lexical-category task. \textbf{b,} Held-out performance of a newly fitted task coordinate relative to 200 label shuffles; this is not transfer of the relation-graph numerical axis. \textbf{c,} Chord-alignment gains for trained Qwen and three full random initializations of the same architecture. \textbf{d,} Cross-fitted relation between internal trajectory displacement and candidate-boundary crossing under the original action budget. Candidate crossing is not answer correctness.}
\label{fig:16}
\end{figure}
\FloatBarrier

\subsubsection*{Can a bounded actuator cross a declared emitted-token boundary selectively?}
We expanded the Qwen action library while fitting the separation coordinate, transition models and boundary objective on training graph groups only. The boundary-trained controller changed the full-vocabulary top-1 output from conflict to clean in 19 of 87 held-out closure prompts. A predeclared confidence-and-entropy gate retained 17 changes while leaving all 203 non-closure top-1 outputs unchanged (Fig. 17e). Crossings occurred for closure override and exception override, but not closure update. Fifty safety-matched action-vector reassignments preserved the gate, action budget and condition-by-layer dose distributions and produced 15--20 changes (mean 17.64; empirical $P=0.804$). The evidence therefore supports selective gating and a bounded local actuator family, not uniquely effective per-trajectory dose assignment. Clean denotes the unperturbed-chain candidate and is generally not the rule-consistent answer under closure; this is emitted-token candidate control, not accuracy improvement.

\begin{figure}[!htbp]
\centering
\includegraphics[width=0.98\textwidth,height=0.78\textheight,keepaspectratio]{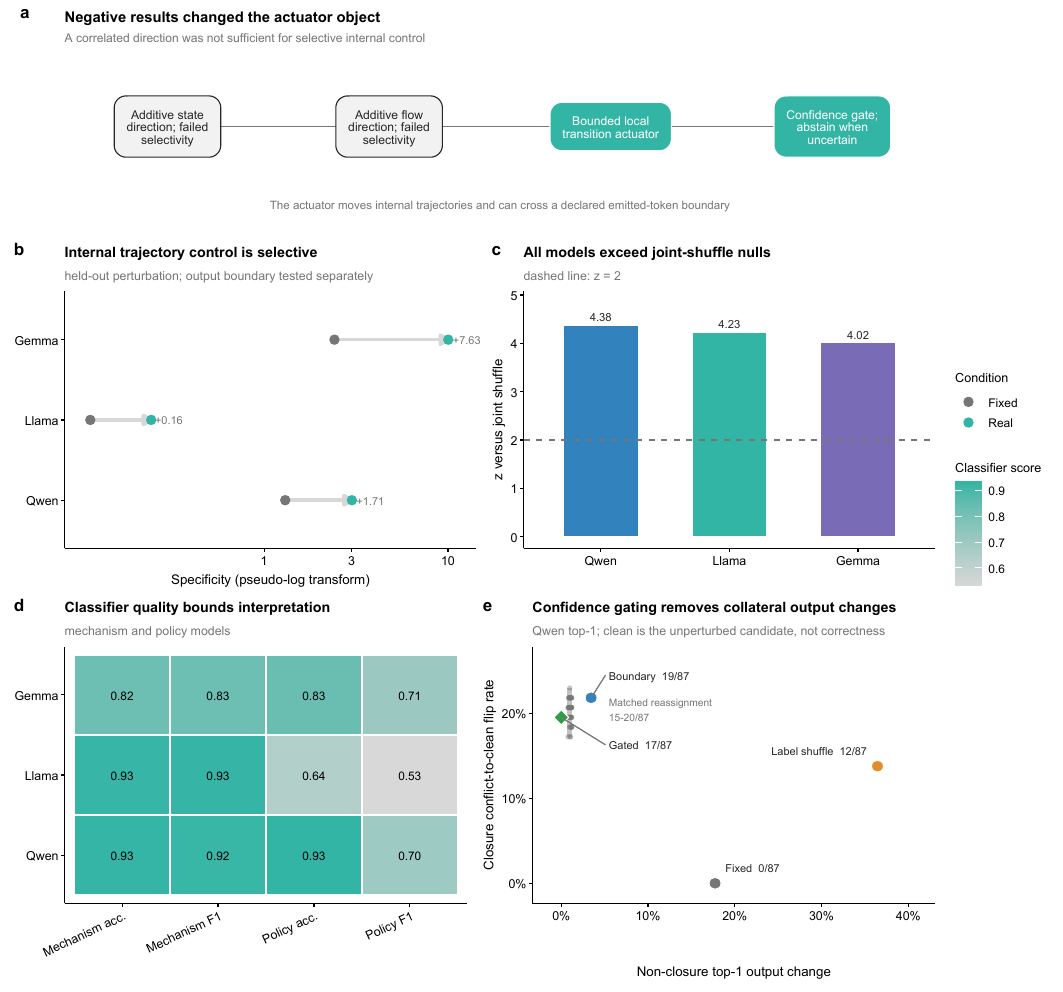}
\caption{A bounded local actuator controls internal trajectories and a declared output boundary. \textbf{a,} Additive intervention failures, transition-actuator switch and confidence gate. \textbf{b,} Internal specificity relative to fixed actions. \textbf{c,} Real policy relative to 50 joint-label shuffles. \textbf{d,} Mechanism and action-policy classifier diagnostics. \textbf{e,} Held-out Qwen closure candidate changes and non-closure top-1 changes, together with 50 gate- and action-budget-matched reassignments. The matched null bounds action-assignment specificity; candidate changes do not imply correctness, open-ended generation control or deployment safety.}
\label{fig:17}
\end{figure}
\FloatBarrier

\section{Discussion}
Across the tested checkpoints, layer order was more than representational chronology: its realized transitions formed a measurable internal process, and perturbing those transitions moved the candidate-competition trajectory in all three checkpoints and crossed a declared emitted-token boundary in Qwen. Transformer components specify how inference is implemented; the ordered trajectory supplies the empirical object on which that computation can be compared and perturbed. The evidence chain runs from direct observation and reduced coordinates to held-out perturbation of the same process.

The account draws on mathematical and physical ideas in deliberately restricted forms. Layer states adapt the trajectory view of discrete dynamical systems, with depth as an evolution index. Checkpoint alignment resembles functional registration. $\Delta U$ uses the compression logic of a reaction coordinate, but we do not test the diffusion-optimality, committor or slow-variable properties required of a physical reaction coordinate. Chord alignment, detour and turning angle test finite-path directional efficiency rather than an exact geodesic equation. The actuator switch separates state estimation from local control action, and confidence gating implements abstention when the state estimate is uncertain. These are sources for executable measurements, not proofs that transformer inference is physically identical to the source systems.

The cross-disciplinary implication is a transferable audit, not a mechanism analogy. Where states and transitions can be operationally measured, a field-specific version can ask whether a chosen process object supports a compressive coordinate, whether an order-sensitive null rejects an endpoint-only explanation, and whether a targeted perturbation alters a predeclared transition or behavioural boundary. A conceptually related systems-neuroscience experiment would compare population trajectories with neuron-wise or static-population summaries and then perturb an inferred transition. Transformer depth is a discrete computational index, whereas neural trajectories evolve in physical time; no shared dynamical equation is implied.

The random-initialization result changes the interpretation of chord-directed geometry. Geometry alone does not identify learned reasoning organization: in the tested Qwen architecture it was stronger before training. Learned functional content is instead supported by task-conditioned separation, checkpoint-specific windows and selective perturbation. Initialization controls in Llama and Gemma remain untested, and pairwise cross-checkpoint direction isomorphism is weak.

Several boundaries remain. The relation-graph $\Delta U$ construction is task-specific; a separately fitted lexical coordinate transferred only partially. Functional windows are exploratory and data-selected. Observed continuous transition coordinates provide descriptive closure when they use target-transition information, but pre-transition estimates recover little of that gain. Output-boundary control was tested only in Qwen, did not transfer to closure-update prompts and did not improve answer correctness. Safety-matched reassignment did not support fine-grained action-to-trajectory dose specificity. The absence of non-closure changes in 203 held-out prompts is an observed selectivity result, not a deployment safety guarantee.

A broader LLM-dynamics landscape may include training shaping the representational substrate, prompts conditioning initial states and available directions, and transformer layers realizing a local unfolding. This study directly tests only the last at inference time. It does not map a training-shaped semantic geometry or establish how prompt-conditioned directions are formed. Those questions require new experiments rather than extrapolation from the present trajectory data.

The research process also provides a bounded methodological observation. Timestamped records show that the core coordinate-mapping, geometric and intervention sequence was developed through a human-directed, AI-assisted hypothesis-to-audit loop. This single case has no no-AI counterfactual and therefore does not estimate an AI-caused productivity gain. It nevertheless demonstrates a reproducible division of labour in which AI assistance expanded operational bandwidth across retrieval, code scaffolding, batch planning and consistency checks, while research-object selection, falsification criteria, interpretation and evidentiary authority remained human.
\section{Methods}
\subsection*{Purpose and scope}
These Methods describe the checkpoints, prompt sets, extraction procedures, controls and analyses used in \emph{High-dimensional layer-state trajectories reveal the internal dynamics of transformer inference}. File and experiment identifiers are retained where they are needed to trace a result to its executable source.

The direct empirical scope is inference-stage trajectory dynamics across Qwen, Llama and Gemma. Training-shaped semantic structure, prompt-conditioned initial conditions and local transformer unfolding are discussed as the broader landscape, but the present experiments directly test the inference-stage layer-state trajectory process.

\subsection*{Model checkpoints}
Analyses used representative instruction-tuned checkpoints from the Qwen, Llama and Gemma families. Model weights are not redistributed and remain subject to provider access conditions.

\begin{table}[H]
\centering\scriptsize\sloppy
\begin{tabularx}{\textwidth}{@{}XXXXX@{}}
\toprule
Model family & Checkpoint & Revision/SHA & Licence tag & Context/config note \\
\midrule
Qwen & \path{Qwen/Qwen2.5-1.5B-Instruct} & \path{989aa7980e4cf806f80c7fef2b1adb7bc71aa306} & \path{apache-2.0} & Public config reports \path{max_position_embeddings=32768}; tokenizer config reports \path{model_max_length=131072}. \\
Llama & \path{meta-llama/Llama-3.2-1B-Instruct} & \path{9213176726f574b556790deb65791e0c5aa438b6} & \path{llama3.2} & Hugging Face metadata is public but files are manually gated; local config reports \path{max_position_embeddings=131072}. \\
Gemma & \path{google/gemma-2-2b-it} & \path{299a8560bedf22ed1c72a8a11e7dce4a7f9f51f8} & \path{gemma} & Hugging Face metadata is public but files are manually gated; local config reports \path{max_position_embeddings=8192} and \path{sliding_window=4096}. \\
\bottomrule
\end{tabularx}
\end{table}

Checkpoint metadata was checked on 21 June 2026. The machine-readable records are indexed in the Source Data manifest.

\subsection*{Prompt inventory}
The controlled prompt inventory contains 96 prompts per checkpoint across 12 conditions. It comprises eight complete relation-graph blocks, each containing every condition once. We used the first eight complete blocks in the frozen source order, rather than sampling individual rows, to preserve condition balance and graph-grouped splitting while holding the extraction and null-control budget constant across checkpoints. This fixed-prefix design was not randomized; conclusions therefore concern the controlled subset and are checked across checkpoints rather than treated as population estimates. Prompt-level metadata and condition summaries are indexed in the Source Data manifest. Prompt conditions are experimental labels, not general natural-language task categories.

\subsection*{Hidden-state trajectory extraction}
For the controlled proxy-audit artifacts reused by the projection-selection geometry control, hidden-state tensors were extracted from Hugging Face model outputs with hidden-state return enabled and key-value caching disabled. For each batch item and transformer layer index \(l\), the stored vector was:

\begin{equation*}
H_{m,p,l}=H^{(l+1)}\!\left[b,\,t_{\mathrm{last}},\,:\right].
\end{equation*}

The \(l+1\) indexing excludes the embedding hidden state. The token position was the last non-padding token from the attention mask, with tokenizer padding side set to left. Models were loaded as float16 on CUDA or float32 on CPU; saved hidden-state arrays were cast to float32. Extraction used a maximum input length of 320 tokens, batch size 4 and random seed 42, without applying a chat template.

\begin{table}[H]
\centering\scriptsize\sloppy
\begin{tabularx}{\textwidth}{@{}XXX@{}}
\toprule
Model family & Hidden-state array shape & Analysed layers \\
\midrule
Qwen & \([960,28,1536]\) & 7-25 \\
Llama & \([960,16,2048]\) & 4-14 \\
Gemma & \([960,26,2304]\) & 6-23 \\
\bottomrule
\end{tabularx}
\end{table}

The projection-selection geometry control used this 96-prompt controlled subset from each 960-row artifact. The machine-readable extraction table is documented in the Source Data manifest.

\subsection*{Projection-selection geometry controls}
Projection-selection geometry controls compared real output weights with row-permuted weights and Gaussian projections. Row-permutation controls permuted vocabulary rows. Gaussian controls matched the analysed real projection matrix by per-dimension mean and standard deviation, then rescaled row norms to the real row-norm distribution.

The projection-selection geometry control used five repeats for the row-permuted and Gaussian projection conditions, with random seed 42. Projection preservation was computed as Pearson and Spearman correlation between pairwise cosine-distance vectors in hidden-state space and readback-centre space.

Readbacks are diagnostic observables. These controls do not imply that vocabulary-facing coordinates exhaust the mechanism coordinate system.

\subsection*{Strict-predecision matched-capacity object audit}
Strict-predecision matched-capacity audit used the frozen 960-observation, 80-relation-graph dataset for each checkpoint. Both hidden and vocabulary-readback proxies were restricted to layers before the declared decision window. The vocabulary proxy comprised adjacent TopK-centre shifts over the predecision layers plus their mean, sum, maximum and standard deviation. The hidden proxy was the mean hidden state over the same predecision window. Neither proxy included decision-window states, \(R_l\), \(dR_l\), candidate logits or generated final labels.

We evaluated matched retained dimensions \(d\in\{2,4,6,8\}\). Within each of five relation-graph-grouped training folds, missing values were median-imputed, features standardized and PCA fitted with exactly \(d\) components; the fitted transforms were then applied to held-out relation graphs. Ridge regression with \(\alpha=10\) predicted \(\Delta U\), and class-weighted logistic regression predicted the controlled mechanism labels. Reported \(R^2\), Pearson correlation and macro-F1 were calculated from pooled out-of-fold predictions. Intervals in Supplementary Fig. S13 are percentile 95\% intervals from 500 bootstrap resamples of relation-graph groups. The audit tests temporal proximity and retained dimensional capacity together. It does not make the controlled mechanism labels independent of the prompt-condition design.

\subsubsection*{Parameter-selection status}
Ridge regularization \(\alpha=10\) was fixed across checkpoints, proxy families and folds rather than selected separately to maximize a reported score. It provides common moderate shrinkage after feature standardization; the present analysis does not claim that 10 is an independently optimized value. The principal geometry analyses used \(K=500\) as the declared high-resolution neighbourhood setting, while the associated K scans and robustness panels report the available lower-K comparisons. The fractional decision interval \(0.70\leq l/L\leq0.92\) was fixed before fitting \(\Delta U\) to isolate the late decision region identified by the preceding layer-flow analyses; it was not selected by optimizing \(\Delta U\) prediction. By contrast, functional-window and best-bias summaries were selected from complete scans on the same controlled dataset and are explicitly treated as exploratory rather than independently confirmed optima.

\begin{table}[H]
\centering\scriptsize\sloppy
\begin{tabularx}{\textwidth}{@{}XXX@{}}
\toprule
Setting & Value & Selection status \\
\midrule
Common random seed & 42 & Fixed implementation seed; shuffle replications used declared offsets. \\
Ridge regularization & \(\alpha=10\) & Fixed after standardization; not tuned per checkpoint or proxy. \\
Principal neighbourhood size & \(K=500\) & Declared high-resolution setting; available K sensitivity is reported separately. \\
Matched retained dimensions & 2, 4, 6, 8 & Prespecified capacity audit grid. \\
Main grouped cross-validation & 5 folds & Relation graph was the split unit. \\
Decision interval & 0.70-0.92 relative depth & Fixed before \(\Delta U\) fitting from the preceding late-transition analysis. \\
Matched-capacity uncertainty & 500 graph bootstraps & Applied to pooled out-of-fold predictions. \\
Internal-modulation split & 30\% held-out graphs & Fixed graph-grouped test partition. \\
Intervention null replication & 50 shuffles per family & Pairing, policy-label and joint null families. \\
\bottomrule
\end{tabularx}
\end{table}

\subsubsection*{Control families and the explanations they test}
\begin{table}[H]
\centering\scriptsize\sloppy
\begin{tabularx}{\textwidth}{@{}XXXX@{}}
\toprule
Control & Preserved & Disrupted & Question tested \\
\midrule
Row-permuted output projection & Output-row set and projection scale & Learned token-row identity & Does learned vocabulary identity supply the preserved geometry? \\
Matched Gaussian projection & Projection-selection pipeline and output dimension & Learned output projection & Is preservation specific to the learned projection? \\
Random-K selection & Number of selected rows & Conditional TopK selection & Does conditional selection contribute to preservation? \\
Predecision exclusion & Graph and condition structure & Decision-window features & Does association survive direct target-window removal? \\
Interior-layer order shuffle & States and trajectory endpoints & Natural interior layer order & Does ordered depth contribute geometric organization? \\
Mechanism/policy label shuffles & Action grid and endpoint calculation & Mechanism-action matching & Is internal specificity reproduced by mismatched policies? \\
\bottomrule
\end{tabularx}
\end{table}

\subsection*{\(\Delta U\) and candidate-margin advantage-flow analyses}
Broad-window and predecision advantage-flow analyses used a clean/conflict label readback margin:

\begin{equation*}
R_l=\operatorname{logit}_l(y_{\mathrm{lead}})-\operatorname{logit}_l(y_{\mathrm{comp}}).
\end{equation*}

For each graph, the clean-anchor trajectory was subtracted:

\begin{equation*}
DR_l^{(g,c)}=R_l^{(g,c)}-R_l^{(g,c_{\mathrm{anchor}})}.
\end{equation*}

Decision layers occupied the declared fractional interval \(0.70\leq l/L\leq0.92\):

\begin{equation*}
\mathcal{L}_{\mathrm{dec}}=\{l:\lfloor0.70L\rfloor\leq l\leq\lfloor0.92L\rfloor\}.
\end{equation*}

For graph \(g\), condition \(c\) and permitted decision layer \(l\in\mathcal{L}_{dec}\), we formed the matrix \(D\) with one row per \((g,c)\) observation and one column per decision layer:

\begin{equation*}
D_{(g,c),l}=dR_l^{(g,c)},\qquad l\in\mathcal{L}_{\mathrm{dec}}.
\end{equation*}

The leading-trajectory support was the clean-label logit and the competing-trajectory support was the conflict-label logit; their difference defined \(R_l\). Thus candidate labels entered the construction only as the two declared readback supports. Generated final answers, correctness labels and recorded pair-choice fields did not enter \(D\), PCA fitting or sign orientation. After centring \(D\), we fitted PC1 and defined

\begin{equation*}
\Delta U_{g,c}=\left(D_{(g,c),:}-\bar D\right)^{\!\top}v_1.
\end{equation*}

The sign was oriented so that the stable-condition mean exceeded the closure-condition mean; if it did not, \(v_1\) and all scores were multiplied by \(-1\). \(\Delta U\) is therefore a bounded dominance observable for the declared candidate pair and window, not a correctness variable. Precursor analyses excluded every column in \(\mathcal{L}_{dec}\). The stricter non-R analysis also removed \(R_l\), \(dR_l\), candidate logits, ranks and other candidate-pair-derived predictors, retaining only predecision TopK/VIM neighbourhood-transition features. Relation-graph groups were never split across train and test folds.

Advantage-flow analyses used an observable layerwise advantage-flow proxy:

\begin{equation*}
O_l^{\mathrm{adv}}=dR_{l+1}-dR_l.
\end{equation*}

Window-level \(O_l^{\mathrm{adv}}\) features were evaluated with relation-graph-grouped cross-validation. \(\Delta U\) prediction used standardized ridge regression with \(\alpha=10\). Mechanism classification used standardized logistic regression with balanced class weights and a 1,000-iteration limit. Relation graph was the split unit.

The broad-window analysis included all transitions and model-specific initialization-to-decision windows. The predecision analysis restricted predictors to transitions before the decision window. The non-readback precursor analysis used the same decision-window \(\Delta U\) target but predicted it from predecision TopK/VIM neighbourhood-transition features rather than \(dR_l\) differences.

Functional windows and the best trajectory-geometry windows were selected and summarized within the same controlled dataset. Graph-grouped folds protect prompt groups during model fitting, but they do not make this analysis-selection step independent. We therefore report complete window scans, retain checkpoint-specific multiplicity in Source Data and treat best-window values as exploratory summaries. No claim in this manuscript depends on an independently confirmed universal window.

For each scanned window, the topology objective was \(1-d_{\mathrm{preserve}}\), where \(d_{\mathrm{preserve}}\) is the mean within-condition preservation distance. The mechanism objective was the relation-graph-grouped mechanism-label macro-F1. The downstream objective, introduced only after \(\Delta U\) was defined, was the grouped correlation with \(\Delta U\). The archived overall scan score was

\begin{equation*}
S_{\mathrm{overall}}=0.5S_{\mathrm{topology}}+0.8\max(S_{\mathrm{mechanism}},0)+0.8\max(S_{\mathrm{downstream}},0).
\end{equation*}

Figure 4 reports topology- and mechanism-selected windows available at that stage of the discovery chain; it does not use the later downstream field to select an earlier result.

\(O_l^{\mathrm{adv}}\) is an observable proxy for path-advantage growth. It should not be conflated with the continuous transition coordinate \(O_l^{\mathrm{cont}}\).

\subsection*{Global dominance-band control}
Band-robustness control defined a global \(\Delta U\) band from negative, critical and positive reference means:

\begin{equation*}
U_{-}=\frac{u_{\mathrm{neg}}+u_{\mathrm{crit}}}{2},\qquad
U_{+}=\frac{u_{\mathrm{crit}}+u_{\mathrm{pos}}}{2}.
\end{equation*}

Rows were assigned to below-band, in-band or above-band phases. Figure 6d reports phase accuracy and macro-F1 from this global band midpoint classifier. This is an auxiliary band-control check, not a primary mechanism label.

\subsection*{One-step transition-state closure}
One-step transition-closure audit audited one-step next-layer state-equation closure. The target vector was:

\begin{equation*}
Y_{l+1}=\left(R_{l+1},b_{l+1},s_{l+1},q_{l+1}\right).
\end{equation*}

The regression model used ridge regularization with \(\alpha=10\) and random seed 42 inside a preprocessing pipeline. Numeric predictors were median-imputed and standardized. Categorical predictors were most-frequent-imputed and one-hot encoded. All reported analyses contained at least five relation-graph groups and therefore used five-fold cross-validation grouped by relation graph. A shuffled non-grouped fallback existed for smaller diagnostic inputs, but that branch did not supply any reported score.

The main feature groups were:

\begin{table}[H]
\centering\scriptsize\sloppy
\begin{tabularx}{\textwidth}{@{}XX@{}}
\toprule
Group & Features \\
\midrule
\(R\) & Candidate margin \(R_l\) and rank gap \(q_l\) \\
\(B\) & Boundary distance \(b_l\) \\
\(T\) & Neighbourhood spread \(s_l\) \\
H-shape & Six local trajectory-shape terms encoding previous-step changes, velocity reversals and centre backtracking \\
\(O_l^{\mathrm{cont}}\) & The first ten continuous transition-coordinate components \\
\bottomrule
\end{tabularx}
\end{table}

The reported score is macro \(R^2\):

\begin{equation*}
R^2_{\mathrm{macro}}=\frac{1}{J}\sum_{j=1}^{J}R_j^2(y_j,\hat y_j).
\end{equation*}

The ten components of \(O_l^{\mathrm{cont}}\) are latent transition coordinates derived from transition signatures. They support a local \(F_O\) state-equation description but should not be described as pure pre-transition observables.

\subsection*{Trajectory-geometry and residual analyses}
For each observed trajectory, the layer-order geometry analysis defined the reference path as the straight chord joining its first and last analysed states, sampled at the same number of layer positions. Shuffle controls retained the same state vectors and endpoints while randomly permuting interior layer order within each trajectory; metrics were recomputed for every shuffle. This null tests whether the observed layer order contains more route organization than an order permutation. It does not match local step lengths, smoothness or autocorrelation, so it cannot distinguish depth-continuity effects from a stronger dynamical constraint. We consequently use geodesic-biased only for the reported real-versus-order-shuffle contrasts and do not claim strict equality between the real trajectory and a shortest path. Versioned implementation filenames are retained only in the code manifest.

Figure 13 display metrics are:

\begin{equation*}
G_{\mathrm{align}}=A_{\mathrm{real}}-A_{\mathrm{shuffle}},\qquad
G_{\mathrm{detour}}=\log_2\!\frac{D_{\mathrm{shuffle}}}{D_{\mathrm{real}}},\qquad
G_{\mathrm{curvature}}=\log_2\!\frac{C_{\mathrm{shuffle}}}{C_{\mathrm{real}}}.
\end{equation*}

The projection-residual analysis compares residual-only, projection-only and mixed projection-residual feature groups. Figure 14b,c reports their correlations with \(\Delta U\):

\begin{equation*}
(\rho_{\parallel},\rho_{\perp},\rho_{\mathrm{joint}})
=\operatorname{corr}\!\left((X_{\parallel},X_{\perp},X_{\mathrm{joint}}),\Delta U\right).
\end{equation*}

The parallel-perpendicular proxy analysis defines feature families for reference-aligned and structured off-direction components. Figure 14d,e reports the separate and combined proxy families:

\begin{equation*}
X_{\mathrm{joint}}=\left[X_{\parallel},X_{\perp}\right].
\end{equation*}

Parallel and perpendicular labels are proxy component families. They should not be described as literal orthogonal decomposition in the full hidden-state manifold.

\subsection*{Internal-modulation replication}
The internal-modulation replication used random seed 42, 96 relation graphs, batch size 4 and a 30\% graph-grouped test partition. Layer windows were model-specific:

\begin{table}[H]
\centering\scriptsize\sloppy
\begin{tabularx}{\textwidth}{@{}XXXX@{}}
\toprule
Model family & Operator layers & Precursor layers & Decision layers \\
\midrule
Qwen & 15-19 & 17-19 & 20-25 \\
Llama & 8-11 & 10-12 & 12-15 \\
Gemma & 13-17 & 15-17 & 18-23 \\
\bottomrule
\end{tabularx}
\end{table}

Mechanism classifiers used the estimated signed and absolute dominance coordinates, the estimated boundary coordinate, velocity magnitude and standardized velocity, normalized layer position, a precursor-window indicator, and baseline dominance and boundary values. The random-forest classifier used 250 trees, maximum depth 7, minimum leaf size 6 and balanced-subsample class weighting; complete implementation fields and seeds are retained in the code manifest.

Mechanism labels were fixed by prompt condition before model fitting. The stable class comprised clean, redundant, irrelevant and paraphrase conditions; competition comprised weak-distractor, balanced-competition and direct-conflict conditions; closure comprised closure-update, closure-override and exception-override conditions. These labels describe the controlled intervention families and were not inferred from generated answers.

Policy classifiers used the mechanism features together with class probabilities and predicted indicators for stable, competition and closure states. The random-forest classifier used 200 trees, maximum depth 6, minimum leaf size 8 and balanced-subsample class weighting; implementation field names are retained in the code manifest.

The policy action grid was; all nonzero weights lay between 0.1 and 0.3:

\begin{equation*}
\mathcal{A}=\{(0,0),(0.1,0.1),(0.2,0.2),(0.3,0.3),
(0.3,0),(0,0.3),(0.3,0.2),(0.2,0.3)\}.
\end{equation*}

The fixed same-combination baseline used operator and precursor weights of 0.30.

Policy labels were fitted only on training graphs. Each training prompt was evaluated on the eight-action grid. For closure prompts, the selected label was the action with the largest signed \(\Delta U\) shift. For stable and competition prompts, it was the action with the smallest absolute shift. The mechanism and policy classifiers used the same graph-level training partition; actions were then predicted for held-out graphs. No row from a held-out graph contributed to operator fitting, precursor directions, probes, dose-grid labels or either classifier. Reported classifier accuracy and macro-F1 are training diagnostics and are not presented as held-out generalization estimates.

Specificity was computed as:

\begin{equation*}
S=\Delta U-\Delta U_{\mathrm{base}},\qquad
\operatorname{specificity}=\mathrm{E}(|s|\mid M=\mathrm{closure})
-\mathrm{E}(|s|\mid M\neq\mathrm{closure}).
\end{equation*}

The 50-shuffle replication used three matched null families: mechanism-policy pairing shuffles, policy-label shuffles and joint shuffles. Mechanism-label shuffles used seeds beginning at 1,042, and policy-label shuffles used seeds beginning at 2,042; each replicate incremented the corresponding seed by one. Joint shuffles used a disjoint seed range before applying the same offsets. Exact seed construction is retained in the code manifest.

In the manuscript, this experiment is described as internal trajectory-dominance modulation. The endpoint does not measure task correctness or deployment control.

\subsection*{Independent lexical-category task and random-initialization control}
The independent task comprised 64 lexical-category problems under six matched conditions. Complete problem groups were split into 44 training and 20 held-out groups. Checkpoint-specific functional windows were frozen from the relation-graph analysis. Geometry was compared with 50 endpoint-preserving interior-layer shuffles per checkpoint. A new task-specific candidate-separation coordinate was fitted on training problems only and evaluated on held-out problems against 200 training-label shuffles. This analysis tested whether the measurement procedure transferred; it did not transfer the numerical relation-graph $\Delta U$ axis.

For the initialization control, three Qwen models were instantiated from the same checkpoint configuration without loading trained weights, using seeds 1701, 1702 and 1703. The tokenizer, 96-prompt controlled subset, hidden-state extraction, normalization, functional window and 50-shuffle endpoint-preserving null were held fixed. The control tests whether chord-directed order requires learned checkpoint parameters; it does not separate every architectural operation or generalize beyond this architecture.

\subsection*{Out-of-fold displacement-to-boundary bridge}
For prompt $p$, the margin and trajectory displacements were $\delta M_p=M_{p,\mathrm{policy}}-M_{p,0}$ and $\delta\Delta U_p=\Delta U_{p,\mathrm{policy}}-\Delta U_{p,0}$. Checkpoint- and regime-specific affine ridge models with penalty $10^{-6}$ estimated $\widehat{\delta M}_p=a_{m,r}+b_{m,r}\delta\Delta U_p$. Five-fold grouped cross-validation held out complete relation graphs, and an inner graph cross-fit supplied training predictions for a one-variable logistic calibrator. Neither the margin-shift model nor its probability calibration evaluated a graph used for fitting. Continuous and crossing intervals used 1,000 graph-bootstrap replicates with seed 20260714.

\subsection*{Confidence-gated emitted-token boundary control}
The output-boundary analysis used Qwen and the seed-42 graph split: 67 training graphs containing 670 prompts and 29 held-out graphs containing 290 prompts, including 87 closure prompts. The separation direction, transition operators, precursor directions, mechanism classifier, action targets and boundary policy were fitted on training graphs only. Generated answer text, correctness and held-out outcomes entered neither fitting nor gate construction.

The bounded action library was
\[
\begin{aligned}
\mathcal A=\{&(0,0),(0.3,0.3),(0.6,0.3),(0.6,0.6),(0.9,0.3),\\
&(0.9,0.6),(0.9,0.9),(1.0,0.6),(1.0,0.9),(1.0,1.2)\}.
\end{aligned}
\]
For closure prompts, training utility was candidate-margin displacement plus a 5-unit crossing bonus minus $0.03(\alpha^2+\beta^2)$; for non-closure prompts it was negative absolute margin displacement minus $0.10(\alpha^2+\beta^2)$. The supervised action classifier used the same 15 pre-intervention features as the internal-control analysis.

For intervention layers $L_I$, mean closure posterior and normalized posterior entropy were $\bar\pi_p=|L_I|^{-1}\sum_{l\in L_I}\pi_{p,l}(\mathrm{closure})$ and $\bar H_p=|L_I|^{-1}\sum_{l\in L_I}[-\sum_k\pi_{p,l}(k)\log\pi_{p,l}(k)]/\log 3$. The predeclared gate was $g_p=I[\bar\pi_p\geq0.50\ \land\ \bar H_p\leq0.90]$. Closed gates applied the identity action at every intervention layer. The primary endpoint was a strict full-vocabulary top-1 continuation change from the declared conflict token to the declared clean token. Non-closure top-1 changes and state-norm ratios were damage guards.

Each of 50 safety-matched repeats permuted the complete five-layer action vector without fixed points among gate-open prompts within the same closure condition. The null preserved the gate-open set, gate-closed zero actions, total action budget and every condition-by-layer action multiset. With 40 of 50 repeats equalling or exceeding the real crossing count, the one-sided empirical value was $(1+40)/(1+50)=0.804$.

\subsection*{Software environments}
Python GPU and model analyses used a frozen Conda environment containing Python 3.11.15, PyTorch 2.5.1+cu121, CUDA 12.1 and Transformers 5.9.0. NumPy 2.4.4, pandas 3.0.3, scikit-learn 1.8.0 and SciPy 1.17.1 supplied the numerical and statistical routines. CUDA was available on one NVIDIA GeForce RTX 4080 SUPER GPU.

Figures 1, 2 and 4-15 and Supplementary Figs. S1-S12 were rendered in a frozen Conda environment with R 4.3.3, ggplot2 3.5.2, ggforce 0.5.0, ggnewscale 0.5.2, patchwork 1.3.2, ggrepel 0.9.6 and cowplot 1.2.0. Data handling used dplyr 1.1.4, tidyr 1.3.1 and readr 2.1.5; SVG and raster export used svglite 2.2.1 and ragg 1.5.0. Figure 3 v0.2 was regenerated after restoration of its source script with R 4.6.1 and ggplot2 4.0.3; its session information is retained beside the figure and in the Source Data manifest. Both environments passed their recorded verification scripts.

\subsection*{Repository and code status}
The public release package contains processed Source Data, machine-readable metadata, analysis scripts, figure scripts and environment manifests. Original analysis scripts are retained for provenance, with repository-relative wrappers where workstation-specific paths were present. The Source Data and code archives are identified by their Zenodo DOIs in the Data and Code availability statements and are mirrored in the linked public GitHub reproducibility branch. File-level correspondence is documented in the Source Data and code manifests rather than in the manuscript text.

\subsection*{Claim boundaries}
The interpretation follows seven constraints. \(\Delta U\) is a bounded dominance observable. Vocabulary readbacks are diagnostics. \(O_l^{\mathrm{adv}}\) and \(O_l^{\mathrm{cont}}\) denote different quantities. Low-dimensional plots are orientation aids. The trajectory-geometry analysis supports chord-directed order relative to declared nulls, not an exact geodesic or training-specific law. Internal modulation is established across the three tested checkpoints; emitted-token candidate-boundary control is established only in the held-out Qwen testbed and does not imply answer correctness, open-ended generation control or deployment safety.

\subsection*{AI-assisted manuscript preparation}
AI-assisted tools were used to organize notes, debug code, check terminology and claim boundaries, and prepare R plotting scripts from author-defined specifications. H.Z. set the research questions, designed and ran the experiments, checked numerical results and figures against Source Data, interpreted the findings and decided which claims to retain or withdraw. Y.W., R.S. and W.C. contributed the domain and methodological support described in the Author contributions statement, and all authors reviewed the manuscript. The authors take responsibility for the manuscript. No scientific conclusion was accepted solely from AI output, and no publication figure was produced by an image-generation system.

{\setlength{\parindent}{0pt}
\section{Data availability}
Processed panel-level tables supporting the preprint analyses are archived on Zenodo at \url{https://doi.org/10.5281/zenodo.21317355} and mirrored in the public project repository at \url{https://github.com/zhaoheng1990-debug/Cognitive_Theory_And_LLM_Phasemap}. The v0.51 release adds source data for the independent lexical task, random-initialization control, held-out displacement-to-boundary bridge, Qwen output-boundary experiment and safety-matched reassignment null. The release contains prompt and checkpoint metadata, hidden-state extraction records, random seeds, null specifications, classifier definitions, figure contracts and file-level checksums. Raw hidden-state arrays are large derivatives of third-party checkpoints and are not redistributed; the prompts, revisions and extraction scripts required to regenerate them are provided subject to checkpoint access and licensing constraints. Model weights are not redistributed.

\section{Code availability}
Analysis scripts, deterministic R figure scripts, environment records, smoke tests and portable path templates are archived on Zenodo at \url{https://doi.org/10.5281/zenodo.21317440} and mirrored in the public reproducibility branch. The v0.51 code release adds the independent-task, random-initialization, boundary-bridge, confidence-gated controller and safety-matched-null workflows. Workstation-specific paths are replaced or wrapped with repository-relative arguments while provenance-preserving source files are retained. Full experiment reruns require the listed model checkpoints and any upstream datasets that cannot be redistributed.

\section{Acknowledgements}
I (H.Z.) am deeply grateful to my family. The initial inspiration for this research arose while I was thinking about how to educate my son. I thank my parents, my wife and my son for supporting my decision to step away from my primary professional work and devote sustained time to an independent research interest. My wife also covered the subscription costs for GPT and Gemini used during the project.

We acknowledge GPT (OpenAI), Gemini (Google), Doubao (ByteDance) and Kimi (Moonshot AI) as AI-assisted research companions. During the exploratory process, these systems helped stimulate ideas, assisted with experimental code and debugging, and provided sustained support for iterative investigation. The authors remained responsible for the research questions, experimental design, execution, verification, interpretation and final scientific decisions, as detailed in the AI-assisted manuscript preparation statement.

I (H.Z.) thank Beihang University, my doctoral advisers Youguang Zhang and Weisheng Zhao, and Hao Sheng, who provided comments on the manuscript. I am also grateful to the senior and junior members of my former laboratory community. Although I did not ultimately complete the PhD degree in microelectronics, the rigorous research training and scientific standards I encountered at Beihang shaped the conduct and presentation of this work.

\section{Funding}
This research received no grant or financial support from any public funding agency, commercial organization or not-for-profit institution. All computing, storage, software subscriptions and other research expenses were provided or paid for personally by H.Z. No resources from the authors' employers were used. The listed affiliations identify the authors and do not imply sponsorship or endorsement by their employers.

\section{Author contributions}
H.Z. conceived the research, defined the questions, formulated the hypotheses developed during the investigation, designed and ran the experiments, established the evidential tests and revised interpretations in response to the results. Y.W. contributed artificial-intelligence domain knowledge and methodological guidance. R.S. contributed operations-research and mathematical-method support. W.C. contributed discussion, advice and conceptual inspiration from medical, neuroscience and cognitive-science perspectives. Y.W., R.S. and W.C. participated in their personal capacities as independent research collaborators and did not provide employer data or resources. All authors reviewed and approved the manuscript. AI-assisted operations are disclosed in Methods and do not constitute authorship.

\section{Competing interests}
The authors declare no competing interests.

\section{Additional information}
Supplementary information accompanies this manuscript. Correspondence and requests for materials should be addressed to Heng Zhao (zhaoheng1990@gmail.com).

}
\clearpage
\appendix
\section{Discovery and execution chronology}
Hidden-state trajectories emerged late in the study. The starting difficulty was practical: intermediate activations were available, with no obvious way to inspect their change across layers. TopK vocabulary readbacks supplied a common readout at each layer and gave the first stable observational surface.

That surface soon proved more useful for asking questions than for settling them. Adjacent readbacks could be fitted by measurable transition maps, and their signatures appeared to cluster into a small number of regimes. The resulting boundary-bulk-boundary picture was appealing but too literal. Fine cluster labels tracked depth strongly, whereas only a coarser late transition survived the layer-clock controls. Different checkpoints also placed comparable structure at different depths, which made model-specific functional windows more defensible than a common layer index.

The next difficulty concerned interpretation. A low-rank coordinate, \(\Delta U\), summarized candidate-relative variation within those windows. Predecision controls found related structure earlier in the stack. The source of the readback geometry was still unclear. Learned vocabulary rows seemed a plausible explanation, yet row permutation left preservation largely intact. Matched Gaussian projections often preserved as much or more. The readback remained useful after its learned-row explanation failed.

This failure prompted the direct comparison that changed the object of the paper. Hidden-state and trajectory-shape variables carried more information about later organization than the vocabulary-facing alternatives in the tested grouped analyses. We therefore followed the ordered hidden states themselves. Transition coordinates, \(\Delta U\), geometric summaries and residual features are measurements of this trajectory, not replacements for it.

The choice remains provisional in the scientific sense. A later strict-predecision matched-capacity audit retained the hidden-state advantage for controlled mechanism labels but found that \(\Delta U\) prediction depended on checkpoint and retained dimension. High-dimensional layer-state trajectories therefore remain the organizing empirical object, without a claim that they outperform readbacks on every metric. This qualified conclusion emerged because an earlier explanation failed.

\clearpage
\begingroup\footnotesize
\setlength{\itemsep}{0pt}
\setlength{\parsep}{0pt}

\endgroup
\end{document}